\documentclass[runningheads]{llncs}

\usepackage{eccv}

\usepackage{eccvabbrv}

\usepackage{graphicx}
\usepackage{booktabs}
\usepackage{bm}

\usepackage[pagebackref,breaklinks,colorlinks,citecolor=eccvblue]{hyperref}
\usepackage{orcidlink}

\begin{document}

%\title{Author Guidelines for ECCV Submission} 
\title{StableGarment: Garment-Centric Generation via Stable Diffusion}
\titlerunning{StableGarment}

\author{Rui Wang\inst{1\dag*} \and
Hailong Guo\inst{1\dag} \and
Jiaming Liu\inst{2\dag} \and
Huaxia Li\inst{2} \and
Haibo Zhao\inst{2} \and
Xu Tang\inst{2} \and
Yao Hu\inst{2} \and
Hao Tang\inst{3} \and
Peipei Li\inst{1\ddag}
}


\institute{Beijing University of Posts and Telecommunications \and
Xiaohongshu Inc. \and Carnegie Mellon University \\
\url{https://raywang335.github.io/stablegarment.github.io/}
}

\authorrunning{Wang et al.}

\maketitle

% \renewcommand{\thefootnote}{\fnsymbol{footnote}}
% \footnotetext[1]{Work done during internship at Xiaohongshu Inc.}
% \footnotetext[$\dag$]{Equally contributed.}
% \footnotetext[\ddag$]{Corresponding author.}
% \renewcommand{\thefootnote}{\arabic{footnote}}

\let\thefootnote\relax\footnotetext{$^{\star}$Work done during internship at Xiaohongshu Inc.}
\let\thefootnote\relax\footnotetext{$^{\dag}$Equally contributed.}
\let\thefootnote\relax\footnotetext{$^{\ddag}$Corresponding author.}

\vspace{-1cm}
\begin{figure}[h]
\hsize=\textwidth 
\centering
  \begin{minipage}{1\textwidth}
    \includegraphics[width=1.0\textwidth]{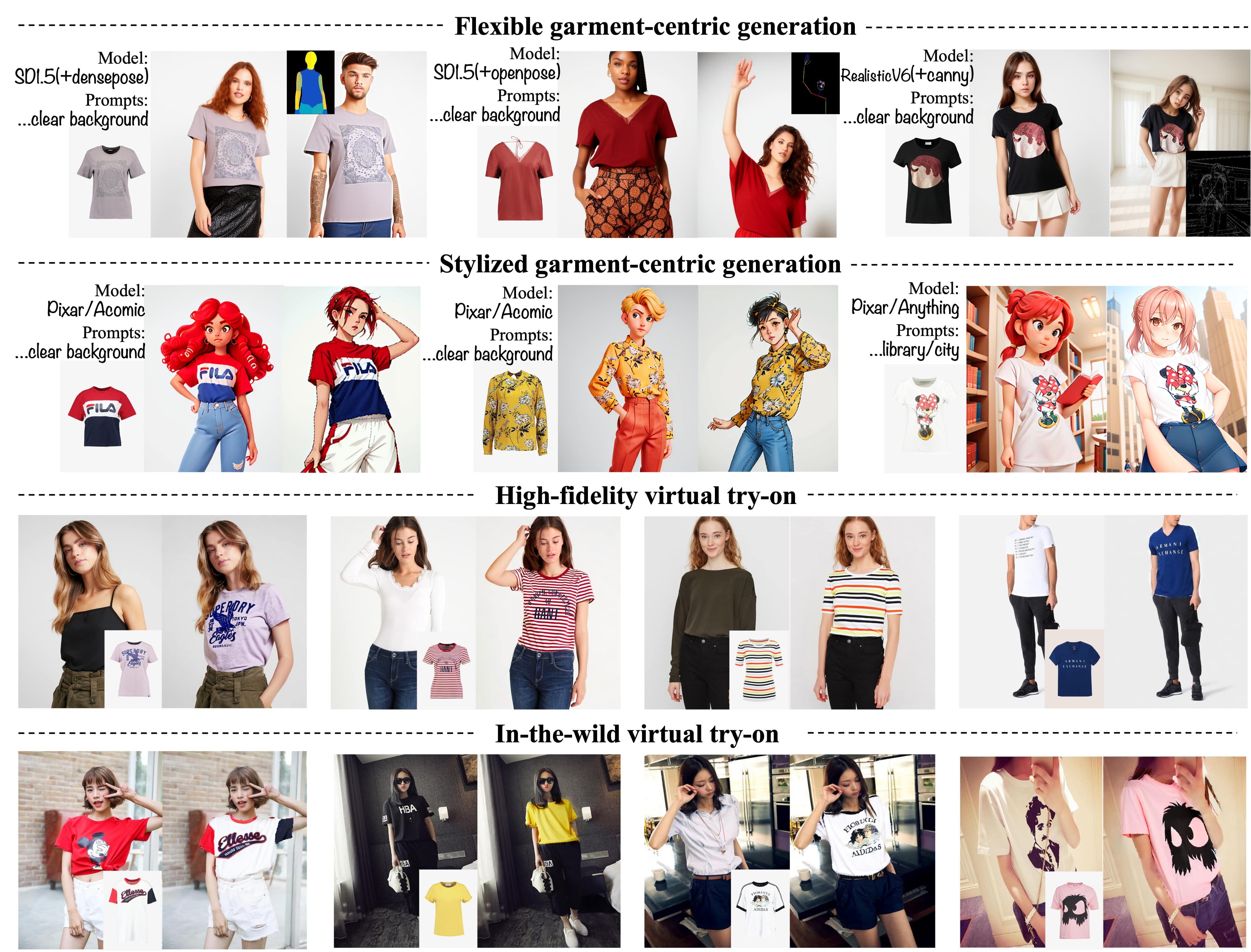}
    \caption{The proposed StableGarment can perform various garment-centric generation tasks. Given a garment input, it could 1) utilize text prompts or control signals to generate a realistic model wearing the garment, 2) support switching stylized models to generate stylized models wearing the garment, and 3) conventional virtual try-on tasks. While performing those tasks, the details of the garment could be well preserved and the garments warping are visually natural.}
    \label{teaser}
    \end{minipage}
    \vspace{-1.3cm}
\end{figure}

\begin{abstract}
In this paper, we introduce StableGarment, a unified framework to tackle garment-centric(GC) generation tasks, including GC text-to-image, controllable GC text-to-image, stylized GC text-to-image, and robust virtual try-on. The main challenge lies in retaining the intricate textures of the garment while maintaining the flexibility of pre-trained Stable Diffusion. Our solution involves the development of a garment encoder, a trainable copy of the denoising UNet equipped with additive self-attention (ASA) layers. These ASA layers are specifically devised to transfer detailed garment textures, also facilitating the integration of stylized base models for the creation of stylized images. Furthermore, the incorporation of a dedicated try-on ControlNet enables StableGarment to execute virtual try-on tasks with precision. We also build a novel data engine that produces high-quality synthesized data to preserve the model’s ability to follow prompts. Extensive experiments demonstrate that our approach delivers state-of-the-art (SOTA) results among existing virtual try-on methods and exhibits high flexibility with broad potential applications in various garment-centric image generation. 
  \vspace{-0.2cm}
\end{abstract}

%To ensure the framework's responsiveness to prompts, we pioneer a novel data engine that produces high-quality synthetic data. Additionally, the incorporation of a dedicated try-on ControlNet enables StableGarment to execute virtual try-on tasks with precision. Through comprehensive experimentation, our framework has demonstrated superior performance, setting new benchmarks in the realm of virtual try-on methodologies. It showcases remarkable versatility, opening avenues for wide-ranging applications across various domains related to garment-centric image generation.

\section{Introduction}
\label{sec:intro}
Recent advances in image generation have seen transformative developments, particularly with the emergence of text-to-image diffusion models trained on large datasets. Among these, Stable Diffusion, an open-source model referenced as \cite{ldm}, stands out for democratizing image generation from textual prompts for a wide user base. Such progress significantly impacts various application domains, notably within the fashion industry, which commands a considerable market presence.

In the realm of fashion, virtual try-on is a classic task that aims to superimpose given garments onto specific user images\cite{zhu2023tryondiffusion,anydoor,dci-vton,dresscode,gp-vton,stableviton,zhu2023tryondiffusion}. The development of diffusion models offers new levels of photorealism in generated images that were previously unattainable with Generative Adversarial Network (GAN)-based methods. Diffusion-based models not only achieve levels of realism previously deemed unattainable, but also excel in restoring intricate details and ensuring the images retain a natural appearance. 

However, when extended beyond conventional virtual try-on tasks, existing methods face notable limitations. Garment merchants are in pursuit of creating varied product visuals, such as posters and display images, more cost-effectively. There is a dual demand: the ability for quick adjustments to models, poses, atmospheres, and backgrounds through textual prompts or reference conditions and the necessity for accurate depiction of textures and fabric dynamics. Stable diffusion's adaptability for swift modifications presents a promising avenue. Recent advances utilizing stable diffusion models in virtual try-on\cite{stableviton} signal the potential for generating garment images via stable diffusion. However, prior works have not fully exploited its capabilities in text-to-image and stylized image creation and have failed to preserve the complete patterns, e.g., stripes and texts.

Therefore, merging the detailed representation of target garments with the adaptable nature of stable diffusion promises to benefit a broader spectrum of users, including merchants, consumers, and artists, by reducing the costs related to garment-related creativity and boosting commercial effectiveness. The question arises: How can we generate images from text prompts or control conditions while preserving the intricate details of specified garments? We address this question by introducing the concept of Garment-Centric (GC) Generation, which focuses on maintaining the fidelity of garment details while enabling flexibility in image creation.

To deal with this problem, we introduce StableGarment, a unified framework built upon Stable Diffusion. This framework is meticulously designed to release the full potential of Stable Diffusion. A garment encoder is devised to encode the details of the target garment. This encoder interfaces with the stable diffusion denoising UNet through an innovative additive self-attention(ASA) mechanism, enhancing the system's versatility in text prompting and model switching. This approach to self-attention facilitates model adaptability for creative try-on purposes. To empower the model with virtual try-on ability, a try-on controlnet is trained. It takes the input of user poses and image contexts and superimposes the garments onto the input image. Moreover, our restructured training dataset, enriched with varied text prompts, enhances the prompt following of the generated images.

To summarize, our contributions are threefold:
\begin{enumerate}
\item We propose a unified framework to address garment-centric (GC) generation tasks, encompassing GC text-to-image, controllable GC text-to-image, stylized GC text-to-image, and virtual try-ons within a single model.
\item We introduce an additive self-attention layer that allows for seamless model switching to stylized base-models and propose a data engine to enhance the models' ability to follow prompts.
\item Our model's performance is benchmarked against existing standards, where it demonstrates state-of-the-art performance among all competitors, underscoring the superiority of our approach.
\end{enumerate}

\section{Related Work}

% \noindent\textbf{Subject-driven Generation.}

\noindent\textbf{Subject-driven Generation.}
Subject-driven generation aims to generate the target subject with text prompts from given reference images. It can be classified into two main categories: test-time finetuning methods~\cite{textureinversion, dreambooth, customdiffusion} and finetuning-free methods~\cite{ip-adapter, blipdiffusion, ssr-encoder,elite,instantbooth}. Finetuning-free approaches offer greater flexibility and are more promising for real-world applications. Typically, finetuning-free methods encode reference images into embeddings or image prompts without requiring additional finetuning. ELITE \cite{elite} proposes global and local mapping schemes, but suffers from limited fidelity. Instantbooth \cite{instantbooth} employs an adapter structure trained on domain-specific data for subject-driven generation without finetuning. IP-Adapter \cite{ip-adapter} encodes images into prompts, while BLIP-Diffusion \cite{blipdiffusion} enables efficient zero-shot setups. However, these methods focus on general subject learning and primarily learn subject representations through the cross-attention layer, which can result in a significant loss of fine-grained details in the generated images. To address this limitation, we propose a garment encoder that captures multi-scale garment features, enabling the generation of results with enhanced fine-grained details.

\noindent\textbf{Stable Diffusion.} Stable Diffusion is a powerful text-to-image model derived from Latent Diffusion Models (LDM)~\cite{ldm}. Its ability to generate high-quality images and perform flexible text editing has contributed significantly to various vision generation tasks. However, there are two significant challenges to achieving a satisfactory generation: (1) controllability and (2) personality.
In the context of controllability, ControlNet~\cite{controlnet} is one of the most effective methods for controlling Stable Diffusion. It consists of a trainable copy of the UNet's encoder with zero initialization and provides structural guidance during image generation.
In terms of personality, IP-Adapter~\cite{ip-adapter} is the first method to utilize image prompts for personality learning and has shown promising results in maintaining semantic consistency. However, when dealing with subjects that contain complicated patterns or text, these methods fail to maintain fidelity. To address this issue, reference-only~\cite{reference-only} methods have been proposed to enhance the details of the learning subject and demonstrate more convincing results. Reference-only nets often consist of a complete copy of UNet, which helps them capture multi-scale texture features across the UNet's feature spaces. Based on these observations, we choose to fully utilize these techniques to solve garment-centric generation tasks.

\noindent \textbf{Virtual Try-on.}
Virtual try-on approaches can be categorized into 3D-based~\cite{aggarwal2022layered, halimi2022pattern, korosteleva2022neuraltailor, majithia2022robust} and image-based methods~\cite{viton, hr-viton, stableviton, ladi-vton, bai2022single, fenocchi2022dual, han2018viton, issenhuth2020not, dresscode, wang2018toward}. Image-based methods are more promising because of their lightweight nature and the ability to generate reasonable results using large-scale try-on datasets.  Image-based virtual try-on methods mainly consist of two stages: warping and blending. Warping methods such as Thin Plate Spline (TPS)\cite{duchon1977splines, viton} and flow-based approaches\cite{viton-hd,gp-vton,bai2022single,dci-vton} have limitations in handling complex deformations and lack structural information about the garment~\cite{chen2023size}. Segmentation maps and DensePose are often used as additional conditions to compensate for information loss caused by occlusion~\cite{hr-viton,viton-hd,li2021toward}. 

Diffusion Probabilistic Models (DPMs) have recently shown promising results in various image synthesis and editing tasks, such as text-to-image synthesis~\cite{ssr-encoder}, image-to-image translation~\cite{pada}, and image inpainting~\cite{lugmayr2022repaint}. In the context of virtual try-on, researchers have explored the use of DPMs to address three key challenges: garment-agnostic preservation, garment detail recovery, and warping accuracy. To preserve garment-agnostic information, many inpainting-based methods~\cite{paint-by-example, anydoor} have been proposed to learn the implicit warping process by inpainting the missing garment region. These methods can learn reasonable warping results but often fail to preserve fine details, such as logos and text, on the garment. To recover garment details, several approaches have adopted advanced techniques, including ControlNet~\cite{controlnet} and image prompt learning, similar to IP-Adapter~\cite{ip-adapter}. DCI-VTON~\cite{dci-vton} combines a diffusion model with a warping module, while LaDI-VTON~\cite{ladi-vton} generates pseudo-words that represent garments, inspired by textual inversion~\cite{gal2022image}. TryOnDiffusion~\cite{zhu2023tryondiffusion} employs cascaded diffusion models with conditions like human pose and garment pose, and StableVITON~\cite{stableviton} leverages a reference garment ControlNet to enhance garment details. WarpDiffusion~\cite{li2023warpdiffusion} incorporates warping garments into diffusion models using local cross attention. In this work, we utilize the garment encoder to recover garment textures and a try-on ControlNet to learn the implicit warping process, effectively addressing the challenges faced by previous methods.

\section{Proposed Method}

\begin{figure*}[!t]
    \centering
\includegraphics[width=1.0\linewidth]{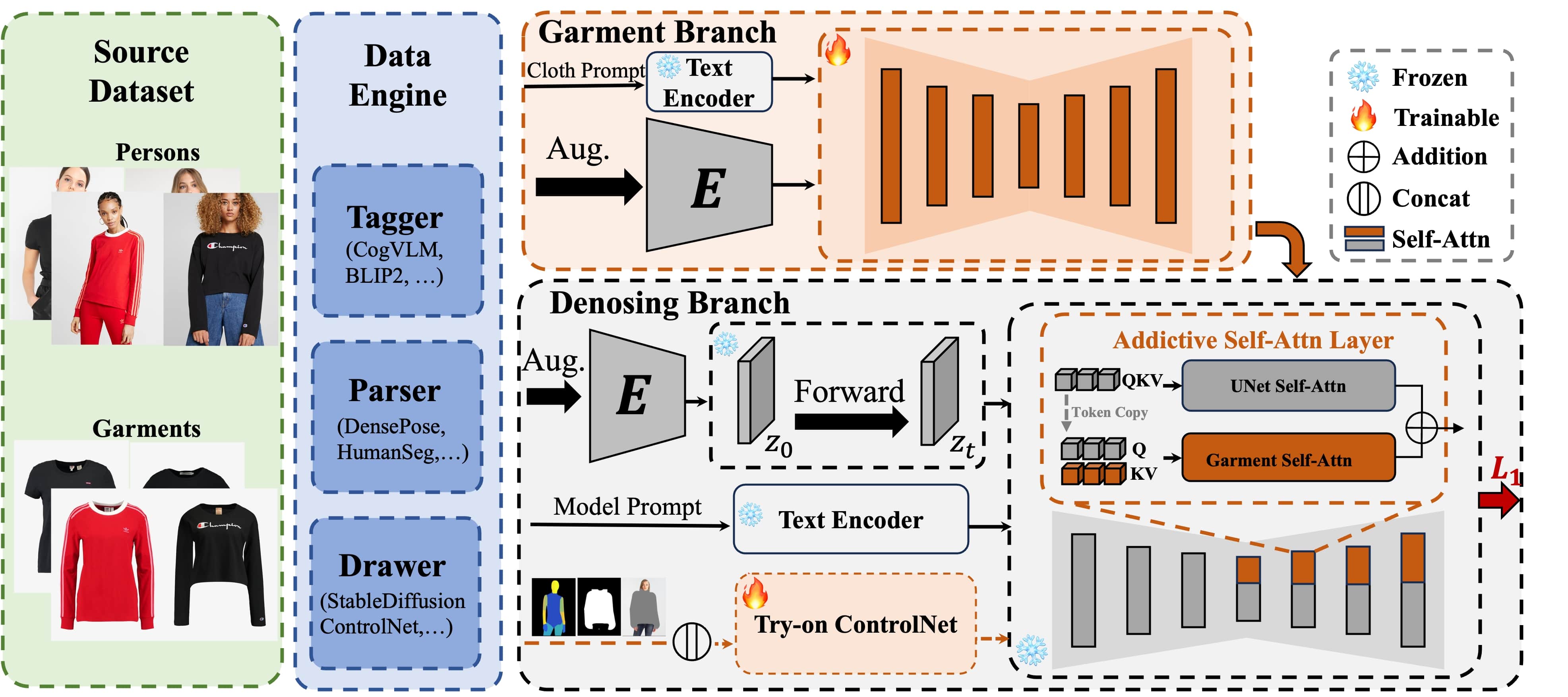}
    \caption{Overview of our framework, consisting of a data engine, garment encoder, and try-on ControlNet. The data engine preserves the model's capacity to follow prompts, while the garment encoder with addictive self-attention layer captures garment details. Meanwhile, the try-on ControlNet is designed for virtual try-on tasks.}
    \label{overview}
    \vspace{-0.6cm}
\end{figure*}

%In this part, we first discuss preliminaries about diffusion models in Section~\ref{sec:Preliminary}, and then introduce the design of our StableGarment in Section~\ref{sec:Architecture}, where we'll reveal the overall architecture, the design of garment encoder, and the design of try-on controlnet. Finally, the specifics of our dataset, training and inference process are detailed in Section~\ref{sec:Training}.

In this section, we begin by exploring the preliminaries on diffusion models as detailed in Section~\ref{sec:Preliminary}. Subsequently, we delve into the architecture of StableGarment in Section~\ref{sec:Architecture}, outlining its overall structure, the details of the garment encoder, and the functionality of the try-on control network. The final part, described in Section~\ref{sec:Training}, provides an in-depth explanation of our dataset along with the training and inference methodologies employed.

\subsection{Preliminary}
\label{sec:Preliminary}
\noindent\textbf{Diffusion Model (DM).}
DM~\cite{ddpm, score-based} belongs to the category of generative models that denoise from a Gaussian prior $\mathbf{x_T}$ to target data distribution $\mathbf{x_0}$ by means of an iterative denoising procedure. The common loss used in DM is:
\begin{equation}
    L_{simple}(\bm{\theta}) := \mathbb{E}_{\mathbf{x_0}, t, \bm{\epsilon}}\left[\left\|\bm{\epsilon}-\bm{\epsilon_\theta}\left(\mathbf{x_t}, t\right)\right\|_2^2\right],
\end{equation}
where $\mathbf{x_t}$ is an noisy image constructed by adding noise $\bm{\epsilon} \in \mathcal{N}(\mathbf{0},\mathbf{1})$  to the natural image $\mathbf{x_0}$ and the network $\bm{\epsilon_\theta(\cdot)}$ is trained to predict the added noise. At inference time, data samples can be generated from Gaussian noise $\bm{\epsilon} \in \mathcal{N}(\mathbf{0},\mathbf{1})$ using the predicted noise $\bm{\epsilon_\theta}(\mathbf{x_t}, t)$ at each timestep $t$ with samplers like DDPM~\cite{ddpm} or DDIM~\cite{ddim}.

\noindent\textbf{Latent Diffusion Model (LDM).}
LDM~\cite{ldm} is proposed to model image representations in autoencoder’s latent space. LDM significantly speeds up the sampling process and facilitates text-to-image generation by incorporating additional text conditions. The LDM loss is:
\begin{equation}
    L_{LDM}(\bm{\theta}) := \mathbb{E}_{z_0, t, \bm{\epsilon}}\left[\left\|\bm{\epsilon}-\bm{\epsilon_\theta}\left(z_t, t, \bm{\tau_{\theta}}(\mathbf{c_t})\right)\right\|_2^2\right],
\end{equation}
where $z_0$ represents image latents and $\bm{\tau_\theta(\cdot)}$ refers to the BERT text encoder~\cite{bert} used to encodes text description $\mathbf{c_t}$.

\noindent\textbf{Stable Diffusion (SD).}
SD is a widely adopted text-to-image diffusion model based on LDM. Compared to LDM, SD is trained on the large LAION~\cite{laion5b} dataset and replaces BERT with the pre-trained CLIP~\cite{clip} text encoder.

\subsection{Network Architecture}
\label{sec:Architecture}

%\subsubsection{Overview.} The overall architecture of StableGarment can be found in Fig.\ref{fig:overview}. Our model consists of three parts, a texture encoder, a dual controlnet, and a pretrained stable diffusion model. We fix the pretrained stable diffusion's parameter, including both VAE and UNet. On top of the UNet, we build a trainable copy of the UNet to take garment as inputs. The texture encoder is enforced with a reference pose branch, which is used for implicit alignment between the in-cloth image and the target pose. There's also a dual controlnet that takes input from pose image and image content. Both the textual encoder and controlnets are connected with the UNet via an additive manner, enabling flexible model switching. 
\noindent\textbf{Overview.} The overall architecture of StableGarment is depicted in Fig. \ref{overview}. The model is segmented into three primary components: a garment encoder, a try-on ControlNet, and a pre-trained stable diffusion model. The parameters of the pre-trained stable diffusion, including the VAE and UNet, are fixed. A trainable duplicate of the UNet is developed on top of the original, specifically designed to process garment input. Additionally, a try-on ControlNet accepts input from both the pose image and the image context. Both the garment encoder and the try-on ControlNet are integrated with the denoising UNet in an additive fashion, thereby offering the flexibility for seamless model switching.

\noindent\textbf{Garment Encoder.} The garment encoder, integral to our model, processes the garment input with a critical objective: to capture the intricate details of the garment. Drawing inspiration from recent studies \cite{reference-only, ip-adapter, controlnet, hu2023animate}, we have devised our encoder composed of garment UNet and a pretrained VAE encoder. The garment UNet is a trainable copy of the denoising UNet and interacts with the denoising UNet via additive self-attention (ASA) layers. The underlying rationale for this configuration is to infuse reference information across multiple scales, thereby extracting the finest details from the reference image while simultaneously preserving its capacity for model switching.

The additive self-attention mechanism is specially designed to allow the encoder integration into various stylized base models. Although numerous studies have explored self-attention for connecting denoising UNet with a reference UNet, these implementations typically adopt a concatenation self-attention(CSA) machanism, as suggested by the reference-only ControlNet\cite{reference-only}. However, our observations indicate that CSA often results in noticeable degradation during base-model switching under complex prompts. Inspired by the design principles of ControlNet\cite{controlnet} and IP-Adapter\cite{ip-adapter}, which both retain the capacity for prompt following and model switching, we have redesigned the connection strategy in an additive format. This revision significantly enhances the system's flexibility and effectiveness in model switching.

Formally, given a reference garment image $I_G$ as input, it first undergoes the VAE encoder $\mathit{E}$ to generate the garment latent, and then the garment UNet $E_g$ generates multi-layered key-value pairs $\{(K_n, V_n)_{n=1}^N\}$, where $n$ denotes the index of self-attention layer of the UNet structure. The $K$s and $V$s are then queried by the $Q^u_n$, the query tensor of the corresponding $n$-th self-attention layer of the denoising UNet. The output value $\mathit{S^u_n}$ of the $n$-th self-attention layer of the denoising UNet is formulated as Eq. \eqref{Equ:Add}.

% \begin{equation}
% %\{(K_n, V_n)_{n=1}^N\}=E_g(I_G, E_c(P_G))
% \{(K_n, V_n)_{n=1}^N\}=E_g(\varepsilon(I_G))
% \end{equation}
% \begin{equation}
% \label{Equ:Add}
% \mathit{S^u_n}
% =\operatorname{Attntion}\left(\mathbf{Q^u_n}, \mathbf{K^u_n}, \mathbf{V^u_n}\right)
% +\operatorname{Attntion}\left(\mathbf{Q^u_n}, \mathbf{K^g_n}, \mathbf{V^g_n}\right),
% \end{equation}
%\begin{equation}
%\begin{aligned}
%\{(K_n, V_n)_{n=1}^N\} &=E_g(\varepsilon(I_G)), \\
%\mathit{S^u_n} &=\operatorname{Attntion}\left(\mathbf{Q^u_n}, \mathbf{K^u_n}, \mathbf{V^u_n}\right)
%+\operatorname{Attntion}\left(\mathbf{Q^u_n}, \mathbf{K^g_n}, %\mathbf{V^g_n}\right),
%\label{Equ:Add}
%\end{aligned}
%\end{equation}
\begin{equation}
\begin{aligned}
\{(K^g_n, V^g_n)_{n=1}^N\} = &E_g(\mathit{E}(I_G)), \\
S^u_n = \operatorname{Attention}\left(Q^u_n, K^u_n, V^u_n\right)&+ \operatorname{Attention}\left(Q^u_n, K^g_n, V^g_n\right),
\end{aligned}
\label{Equ:Add}
\end{equation}
where $\operatorname{Attention}(Q,K,V)$ is the attention operator that calculate attention via query $Q$, key $K$, and value $V$. 

\noindent\textbf{Try-on ControlNet.}
To enable the StableGarment framework for virtual try-on tasks, we have developed a try-on ControlNet. This network is engineered to integrate the target body shape and image context into the denoising UNet workflow. The target body shape is precisely captured using DensePose\cite{detectron2}, while the image context includes elements such as the background, skin tone, and attributes. A garment mask $m$ is employed to highlight areas in need of inpainting. Acknowledging the tight correlation between image context, inpainting area, and the pose, these conditions are concatenated into a unified input to the try-on Controlnet. Starting with an image context $I$, a garment mask $m$ is produced following the methodology outlined in \cite{viton-hd}. Consequently, the inputs for the try-on ControlNet comprise the garment mask $m$, the masked image $I_m=I\cdot m$, alongside pose guidance $p$. The control signals of the ControlNet are directly infused into the denoising UNet as feature residuals. This setup enables the try-on ControlNet to deliver precise, pixel-aligned control signals for virtual try-ons, drawing from the provided image context to ensure accurate garment alignment.

Formally, the try-on ControlNet $E_c$ processes the three aforementioned conditions as in Eq. \eqref{Equ:Contorlnet}, and the residuals of the features are added to the denoising UNet.
\begin{equation}
\label{Equ:Contorlnet}
\mathit{R^u}=E_c(I_m \odot m \odot p), 
\end{equation}
where $R^u$ represents feature residuals of try-on ControlNet, and $\odot$ denotes the concatenation operation. 

\noindent\textbf{Text Prompt.}
To preserve the model's ability to follow prompts, we dispatch distinct prompts to both the garment UNet and the denoising UNet. Specifically, the garment UNet receives prompts of the garment category, facilitating its understanding of the input garment's dynamic nature. Concurrently, the denoising UNet is prompted with descriptions of the target image, maintaining its proficiency in generating images responsive to a diverse range of textual cues.

\subsection{Training Strategy and Inference}
\label{sec:Training}
\noindent\textbf{Dataset Preparation.}
For the preparation of our dataset, we employ a pre-trained model to generate high-quality synthesized garment data, which is crucial for a precise and controllable garment-centric learning process. The primary objective of the data engine is to generate garment-centric images across a wide range of text prompts, thereby preserving the model's ability to follow text prompts throughout the learning phase. A detailed demonstration can be found in the supplementary material. Our data engine is divided into three modules:
\begin{itemize}
    \item \textbf{Parser:} The parser module is tasked with creating a parse map and dense map from an input image. It incorporates a pre-trained parsing segmentation network alongside Detectron2~\cite{detectron2}, facilitating accurate segmentation of the garment.
    %\item \textbf{Tagger.} The tagger is designed to generate an accurate and complete text description given a target garment image. It employs CogVLM-chat~\cite{cogvlm} to generate text label and GPT4 to generate inpainting templates for the drawer module.
    \item \textbf{Tagger:} The tagger module generates detailed text descriptions for target garment images. It utilizes CogVLM-chat~\cite{cogvlm} for generating text descriptions, and GPT4 to create inpainting templates, which are subsequently used by the drawer module.
    \item \textbf{Drawer:} The drawer module functions as the synthesized data generator. It uses ControlNet~\cite{controlnet} in conjunction with a pre-trained inpainting model, epiCRealism~\cite{epicinapint}, to produce the final synthetic garment images.
\end{itemize}

\noindent\textbf{Learning Objectives.} The training for the StableGarment model incorporates a two-stage strategy. Initially, in the first phase, we utilize synthetic data to train our garment encoder. This training focuses on accurately capturing the nuances of clothing details while maintaining the ability for textual modifications. In the second stage, we freeze our garment encoder and only train our try-on ControlNet. We persist with the same learning objective, incorporating tryon-specific conditions.
The loss functions for the two stages are expressed as Eq. \eqref{equ:loss}:
\begin{equation}
\begin{aligned}
    L &= \mathbb{E}_{z_0, t, I_{g},\epsilon\in\mathcal{N}(0,1)} \left[ ||\epsilon - \epsilon_{\theta}||_2^2\right], \\
    L &= \mathbb{E}_{z_0, t, I_{g}, p, m, I, \epsilon\in\mathcal{N}(0,1)} \left[ ||\epsilon - \epsilon_{\theta}||_2^2\right],
\end{aligned}
\label{equ:loss}
\end{equation}
where $I_{g}$ and $I$ are the garment reference image and the image context respectively, $p$ is the pose guidance from the image context, and $m$ denotes the corresponding garment mask.

\noindent\textbf{Inference.}
As illustrated in Table \ref{tab:tasks}, StableGarment is capable of dealing with various tasks. In addition to using different components for different tasks, the inference pipelines are different. 
In the context of garment-centric generation tasks, our method is flexible and generalizable to other third-party plugins, e.g., stylized base models and ControlNets. Specifically, for virtual try-on tasks, we employ an inpainting pipeline to achieve both virtual try-on consistency and cloth-agnostic region preservation. Concretely, each updating denoising step can be rewritten as:
\begin{equation}
    \epsilon_{t-1}^{\prime} = m \cdot \epsilon_{t-1} + (1-m) \cdot z_{t-1}^{src},
\end{equation}
where $z_{t-1}^{src}$ represents the noised image latents from the source model input and $\epsilon_{t-1}$ represents predicted noise.
This adjustment ensures the seamless preservation of background elements and model details. 

\begin{table}[t]
    \centering
    \caption{The StableGarment is capable of performing various garment-centric(GC) tasks. The configurations for performing different tasks are listed in the table.}
    \vspace{-0.1cm}
    	\resizebox{1\linewidth}{!}{% 
    \begin{tabular}{c|c|c|c|c}
         \toprule
         Task & Pipeline & Garment encoder & Base model & ControlNet  \\
        \midrule
         GC t2i& text2image & $\checkmark$ & stable diffusion v1.5 & $\times$ \\
         stylized GC t2i & text2image & $\checkmark$ & stylized base-model & $\times$ \\
         controllable GC t2i & text2image & $\checkmark$ & any base-model & any ContorlNet \\
         virtual try-on & inpainting & $\checkmark$ & stable diffusion v1.5 & try-on ControlNet \\
         \bottomrule
    \end{tabular}}
    \label{tab:tasks}
        \vspace{-0.2cm}
\end{table}

\section{Experiments}

\noindent\textbf{Baselines.} 
For subject-driven generation, we compare our method with three finetuning-free method, including ELITE~\cite{elite}, IP-Adapter~\cite{ip-adapter} and BLIP-Diffusion~\cite{blipdiffusion}. 
In the context of virtual try-on task, we compare our method with three GAN-based virtual try-on methods, VITON-HD~\cite{viton-hd}, HR-VITON~\cite{hr-viton} and GP-VTON~\cite{gp-vton}, and three Diffusion-based virtual try-on methods, LADI-VTON~\cite{ladi-vton}, DCI-VTON~\cite{dci-vton} and StableVITON~\cite{stableviton}. We also compare our method with Diffusion-based inpainting methods, Paint-by-Example~\cite{paint-by-example} and AnyDoor~\cite{anydoor}. To ensure a fair comparison, we utilize the official implementations for all the aforementioned methods at 512$\times$384 resolution.

\noindent\textbf{Datasets.} 
For subject-driven generation, we introduce a garment-centric generation benchmark derived from a subset of the VITON-HD test set. This benchmark, which includes 14 garments and 6 text prompts, aims to cover a broad spectrum of garment attributes such as color, shape, pattern, size, and type. The prompts, crafted through GPT-4, integrate a diverse range of skin tones, hairstyles, backgrounds, and additional attributes, enabling a robust evaluation against subject-driven generation approaches. The details of proposed benchmark will be provided in the supplementary material. 

In the context of the virtual try-on task, we employ two widely-used, publicly available high-resolution datasets for virtual try-on: VITON-HD~\cite{viton-hd} and Dress Code~\cite{dresscode}.
We keep the same train-test split as the previous methods~\cite{stableviton, ladi-vton} on both of these datasets. In our study, we only use the official datasets for training the virtual try-on tasks to ensure fairness.
% VITON-HD contains 13,679 frontal-view woman and top clothes image pairs at the resolution of 1024×768. Following previous work\cite{viton-hd, stableviton}, we split the dataset into a training set and a test set with 11,647 and 2,032 pairs, respectively, and conduct experiments at two different resolution. DressCode dataset features over 53,000 image pairs of clothing and human models wearing them. The dataset includes high-resolution images (i.e., 1024×768) and garments belonging to different macro-categories, such as upper-body clothes, lower-body clothes, and dresses. In our experiments, we focus on the upper clothes and employ the original splits of the dataset following previous methods~\cite{stableviton, ladi-vton}.

%We also build a garment-centric genearation benchmark. To compare with subject-driven methods, we make a benchmark composed of a subset of VITON-HD test set. The benchmark is consisted of 14 garments and 6 prompts. We selected the benchmark with consideration for covering a comprehensive range of garment categories, including factors such as color, shape, decorative patterns, size, and type. The prompts are generated by GPT-4, incorporating various skin colors, hair styles, backgrounds, and other attributes.

\begin{figure*}[!t]
    \centering
    \vspace{-0.1cm}
\includegraphics[width=1.0\linewidth]{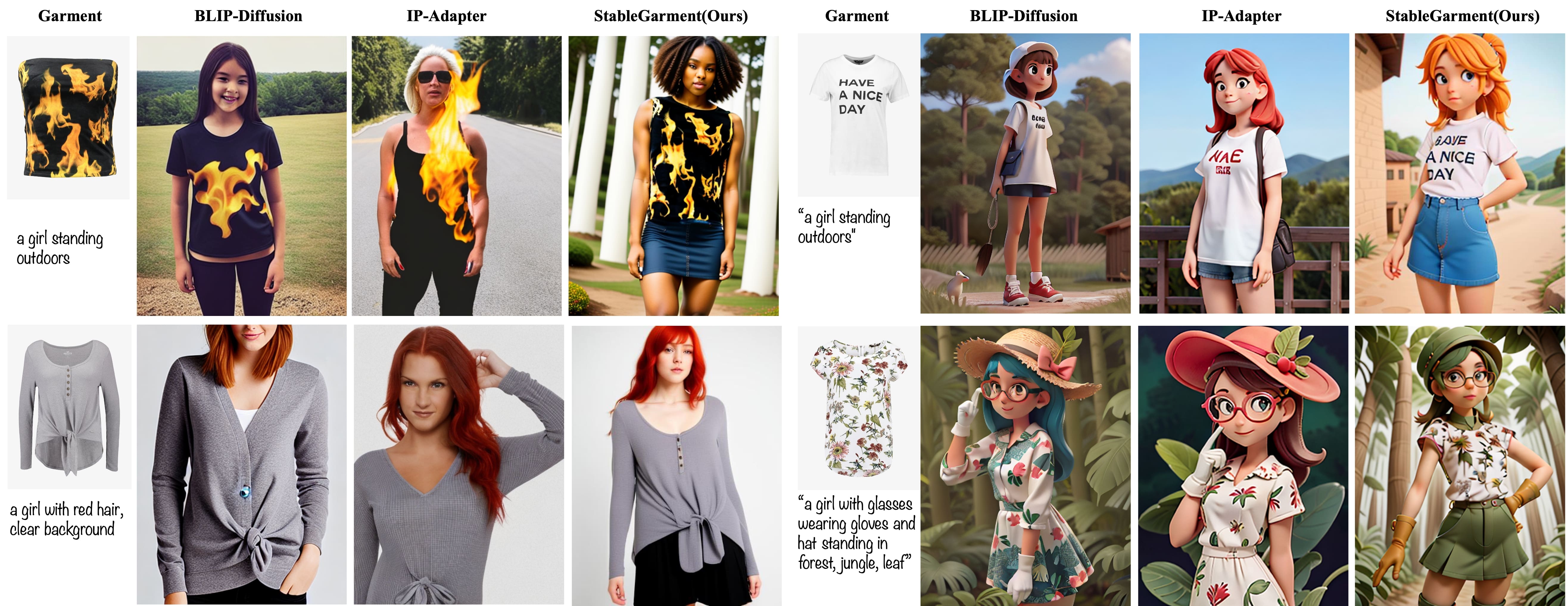}
    \caption{Comparison with subject-driven generation methods.}
    \label{t2i-comp}
        \vspace{-0.6cm}
\end{figure*}

\noindent\textbf{Evaluation Metrics.} %For fairness, we only use the official dataset to train virtual try-on task. Following previous methods, we employ SSIM~\cite{ssim} and LPIPS~\cite{lpips} for paired settings and FID~\cite{fid} and KID~\cite{kid} for unpaired settings. We incorporate human perception by conducting a user study for a more thorough comparison. 
For virtual try-on task, we keep consistent with previous methodologies and apply SSIM~\cite{ssim} and LPIPS~\cite{lpips} metrics for evaluation under paired settings, and FID~\cite{fid} and KID~\cite{kid} metrics for unpaired settings. Following previous studies~\cite{ssr-encoder,avrahami2023break,ruiz2023dreambooth,voynov2023p+}, we utilize the DINO-M metric to assess the fidelity of the generated image to the reference image. The DINO-M metric compares a masked version of the try-on image with the garment image and computes the similarity using image embeddings extracted by the DINO~\cite{caron2021emerging} model, reflecting the identity of the garment. As the objectives of subject-driven generation differ from try-on, we employ three distinct metrics CLIP-T, CLIP-I~\cite{hessel2021clipscore} and Aesthetic score~\cite{laion5b} to compare with subject-driven generation methods. 
%SSIM measures the quality of three key features: luminance, contrast, and structure between the reconstruction results and the ground truth.
%LPIPS evaluates the perceptual similarity. 
%FID and KID is computed for evaluating the generation performance. Following previous studies~\cite{ssr-encoder,avrahami2023break,ruiz2023dreambooth,voynov2023p+}, we utilize the Dino-M metric to assess the fidelity of generated image to the reference image. The Dino-M metric compares a masked version of the try-on image with the reference garment image, and computes the similarity using image embeddings extracted by the DINO model, reflecting the identity of the garment.
%We have gathered composite images created by various methods for 300 randomly chosen pairs from the test set, with a resolution of 512 × 384. We enlisted 20 human evaluators to identify which method best reconstructs the clothing and which yields the most lifelike outcomes for each pair. Subsequently, we tally and present the number of times each method is preferred as the top performer in these two categories.
%Implementation details will be provided in the supplementary material. 

We adopt two metrics, human preference and human scores, to integrate human judgment into our evaluation process. In our experimental setup, we compile composite
images generated by various methods for 100 randomly selected pairs from the test set. Following~\cite{stableviton}, totally 200 evaluators were asked to evaluate the three aspects of the try-on results, garment identity, try-on quality and garment-agnostic preservation. In each aspect, human preference showcases the frequency with which each method was selected as the top choice in these baselines, and human scores reflect the weighted evaluation scores based on their relative ranking order. The formula computation and our implementation details will be provided in the supplementary material.

\subsection{Qualitative Results}

\begin{figure*}[!t]
    \centering
\includegraphics[width=1.0\linewidth]{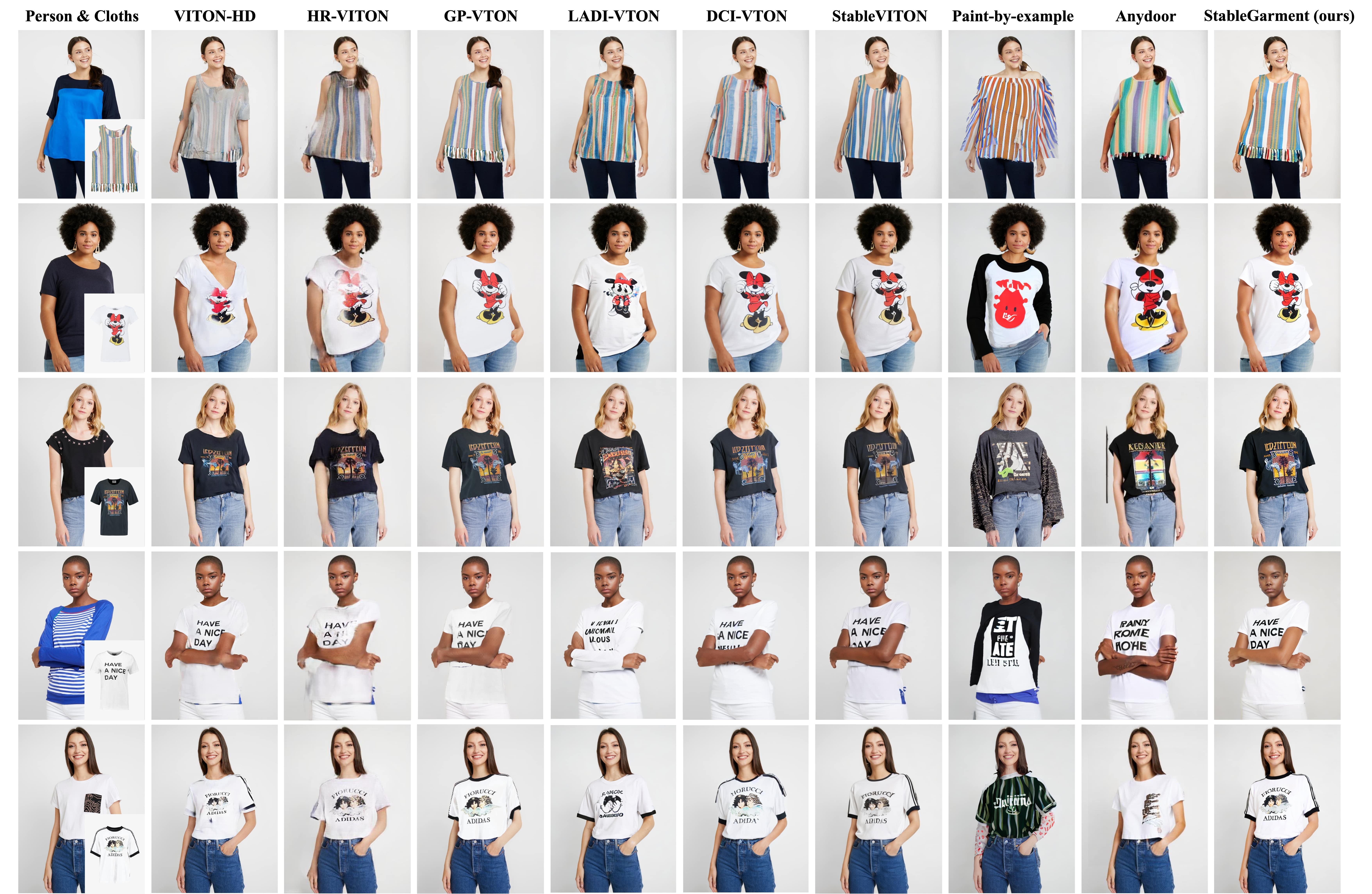}
    \caption{Qualitative comparison with baselines on VITON-HD dataset.}
    \label{viton-hd-comp}
    \vspace{-0.5cm}
    
\end{figure*}

\begin{figure*}[!t]
    \centering
\includegraphics[width=1.0\linewidth]{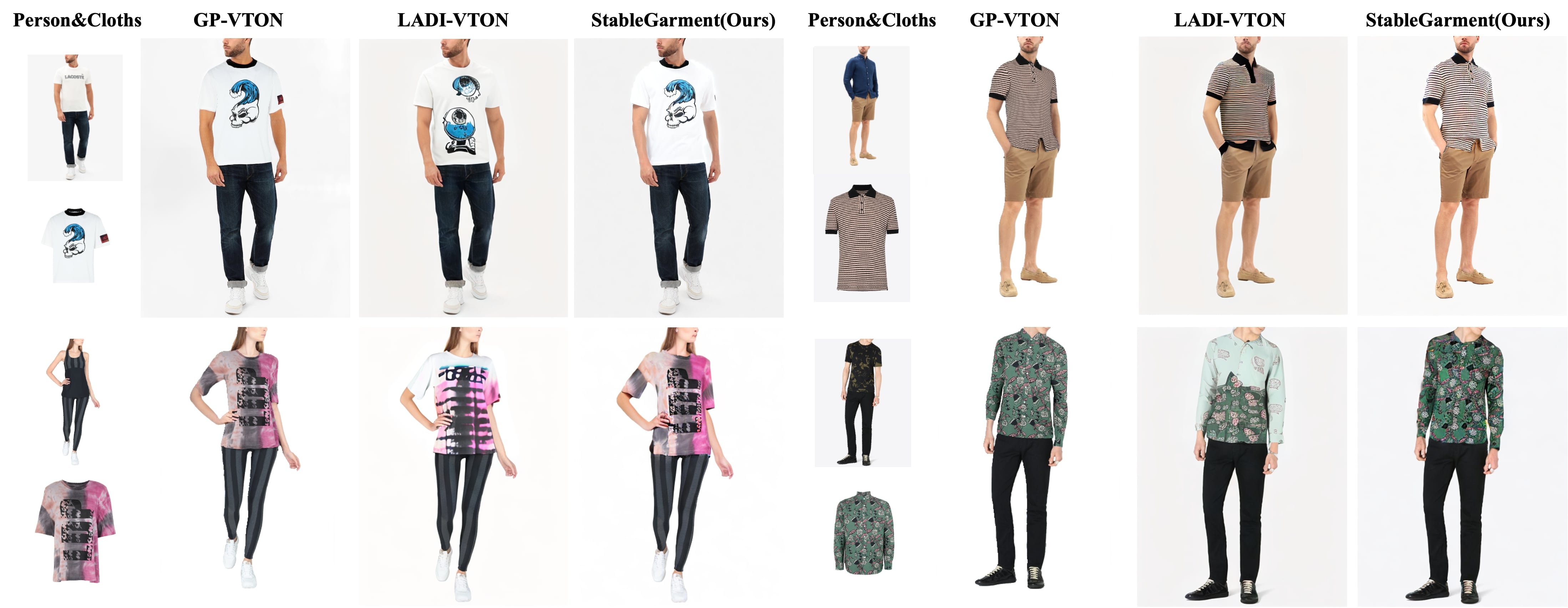}
    \caption{Comparison on Dress Code dataset.}
    \label{dresscode-comp}
        \vspace{-0.6cm}
\end{figure*}

\noindent \textbf{Subject-driven Generation Results.} 
Fig.~\ref{t2i-comp} displays our StableGarment can generate reasonable outputs following the different prompts while preserving clothing texture under different diffusion base models. BLIP-Diffusion and IP-Adapter can only preserve partial texture features of the garment, making them prone to losing details such as shape and text. In contrast, our method not only achieves better fidelity in texture, shape, and detail preservation but also offers adaptability in style transfer when switching base models.

\noindent \textbf{Virtual Try-on Results.} Fig.~\ref{viton-hd-comp} shows a visual comparison of our outputs to other baselines on VITON-HD. Although previous methods synthesize reasonable warping garments and preserve the background well, they tend to generate blurry try-on results with manifest artifacts. Diffusion-based methods greatly improve the garment generation quality, but most of these methods fail to capture accurate clothing texture, e.g. text and tiny patterns. Due to the meaningful feature spaces of our proposed garment encoder, the try-on cloth successfully preserves the slight details. For example, our method accurately captures the `ADIDAS' on the cloths in the last row. The quantitative results on VTION-HD convincingly verify that our method can achieve a more realistic try-on effect for these garments. More examples of our virtual try-on results will be reported in the supplementary materials.
    
We also provide a visual comparison of our outputs to other baselines on the Dress Code in Fig.~\ref{dresscode-comp}. Compared to GP-VTON and LADI-VTON, our try-on results exhibit a more natural appearance.

\subsection{Quantitative Results}

\noindent\textbf{Comparison with Subject-driven Generation Methods.} We propose a garment-centric benchmark as described above to evaluate our method and other baselines, which includes 14 garments and 6 different prompts. Table~\ref{Tab:subject} provides a detailed quantitative evaluation of these methods. Overall, both IP-Adapter and our method demonstrate the ability to generate accurate outputs that adhere to the given text descriptions. Moreover, our method clearly outperforms previous state-of-the-art fine-tuning-free methods on the CLIP-I and Aesthetic score metrics, showcasing the capability of our garment encoder to capture comprehensive and fine-grained features from reference garments.

\begin{table}[!t]
  \small       %此处写字体大小控制命令 
  \centering      %表格居中
  \setlength{\abovecaptionskip}{0cm}
  \caption{Quantitative comparison of subject-driven generation methods. The best and the second best results are denoted as \textbf{Bold} and \underline{underline}, respectively.}         %表标题
  \vspace{-0.2cm}
  \label{Tab:subject}         %标签，用作引用
\begin{tabular}{l|ccc}
  \toprule
  \textbf{Model}                         & CLIP-T$\uparrow$    & CLIP-I$\uparrow$ & AS$\uparrow$ \\
  \midrule
  BLIP-Diffusion~\cite{blipdiffusion}     &  0.187            & \underline{0.692}     &    4.973         \\
  IP-Adapter~\cite{ip-adapter}            &  \textbf{0.274}            & 0.628        &    \underline{5.165}         \\
  ELITE~\cite{elite}                     &  0.244                & 0.617             &    4.468       \\
  Ours                                   &  \underline{0.262}              & \textbf{0.719}                  &   \textbf{5.255}         \\
  \bottomrule
\end{tabular}
\vspace{-0.2cm}
\end{table}

\begin{table}[!t]
  \small       %此处写字体大小控制命令 
  \centering      %表格居中
  \setlength{\abovecaptionskip}{0cm}
  \caption{Quantitative comparison on VITON-HD dataset. We multiply KID by 100 for better comparison. The best and the second best results are denoted as \textbf{Bold} and \underline{underline}, respectively.}         %表标题
  \label{Tab:main}
  \vspace{-0.2cm}
\scalebox{0.95}{%标签，用作引用
\begin{tabular}{l|ccccc}
  \toprule
  \textbf{Model} & \textbf{LPIPS}↓  & \textbf{SSIM}↑ & \textbf{FID}↓ &\textbf{KID}↓ & \textbf{DINO-M}↑  \\
  \midrule
  VITON-HD~\cite{viton-hd}  &  0.117  & 0.862 & 12.12 & 0.323 & 0.615 \\
  HR-VTON~\cite{hr-viton}   &  0.105  & 0.868 & 11.27 & 0.273 & 0.634 \\
  LADI-VTON~\cite{ladi-vton}&  0.092  & 0.875 & 9.37 & 0.158 & 0.633 \\
  DCI-VTON$^*$~\cite{dci-vton}  &  0.092  & 0.876 & 9.44 & 0.164 & 0.648 \\ 
  StableVITON~\cite{stableviton}   &  0.083  & 0.866 & \underline{8.19} & \underline{0.118} & 0.647 \\
  GP-VTON$^*$~\cite{gp-vton}&  \underline{0.083}  & \textbf{0.887} & 9.83 & 0.141 & \textbf{0.675} \\
  Paint-by-Example~\cite{paint-by-example}   &  0.217  & 0.776 & 14.17 & 0.825 & 0.590 \\
  AnyDoor~\cite{anydoor}    &  0.139  & 0.828 & 12.73 & 0.544 & 0.641 \\
  Ours   &  \textbf{0.077}  & \underline{0.877} & \textbf{7.98} & \textbf{0.104} & \underline{0.661} \\
  \bottomrule
\end{tabular}}
\vspace{-0.2cm}
\end{table}

\noindent\textbf{Comparison with Visual Try-on Methods.} We compare our method with existing baselines on the VITON-HD dataset and report the results in Table~\ref{Tab:main}. Our method demonstrates competitive performance across all metrics compared to the baselines, particularly in the unpaired metrics(i.e., FID and KID).  The success of our approach can be attributed to the garment encoder and DensePose-based try-on ControlNet, which effectively capture fine-grained garment features and generate high-quality try-on images. Furthermore, we utilize DINO-M to evaluate the recovery of cloth texture in the unpaired setting. Under this metric, our method displays competitive performance, further demonstrating the superiority of our method in the virtual try-on task.

% We compare our method with existing baselines on the VITON-HD dataset and report the results in Table~\ref{Tab:main}. Our method outperforms all the baselines in both 512$\times$384 and 1024$\times$768 resolution settings, particularly in the unpaired metrics (i.e., FID and KID).  

\begin{table}[!t]
  \small       %此处写字体大小控制命令 
  \centering      %表格居中
  \setlength{\abovecaptionskip}{0cm}
  \caption{User Study on virtual try-on task. We evaluate our methods with other four baselines on three metrics: garment identity, try-on quality and garment-agnostic preservation. The best and the second best results are denoted as \textbf{Bold} and \underline{underline}, respectively.}         %表标题
  \label{Tab:userstudy}      
  \vspace{-0.2cm}
  \scalebox{0.85}{%标签，用作引用
\begin{tabular}{l|ccc | ccc}
  \toprule
  & \multicolumn{3}{c|}{\textbf{Human Preference(\%)$\uparrow$}} & \multicolumn{3}{c}{\textbf{Human Scores$\uparrow$}} \\
  \textbf{Model}                    & Identity          & Quality        & Preservation      & Identity       & Quality         & Preservation\\
  \midrule
  GP-VTON~\cite{gp-vton}            &  \underline{15.35}          & 13.82          &    15.85          & \underline{2.98}          & 2.72            & 2.68    \\
  DCI-VTON~\cite{dci-vton}          &  13.68           & 12.51          &    15.13          & 2.82          & 2.64            & 2.78    \\
  LADI-VTON~\cite{ladi-vton}        &  10.88           & 12.56          &    14.06          & 2.12          & 2.22            & 2.41      \\
  StableVITON~\cite{stableviton}    &  13.40           & \underline{14.32}          &    \underline{22.77}          & 2.69          & \underline{2.99}            & \textbf{3.02}   \\
  Ours                              &  \textbf{46.70}  & \textbf{46.80} &   \textbf{27.30}  & \textbf{3.15} & \textbf{3.12}   & \underline{2.99}    \\
  \bottomrule
\end{tabular}}
\vspace{-0.4cm}
\end{table}

\subsection{User Study}
To further evaluate the generation quality of our model, we conduct a user study to measure both the realism of the generated images and their coherence with the inputs given to the virtual try-on model. As shown in Table \ref{Tab:userstudy}, our method outperforms other baselines in garment texture and try-on quality, demonstrating the effectiveness of our approach in preserving garment texture and maintaining try-on quality. Although our method does not achieve the best performance in the traditional quantitative metrics, it still showcases satisfactory performance in garment-agnostic preservation.

\subsection{Ablation Study}
We take the 512 $\times$ 384 resolution on the VITON-HD dataset as the basic setting and perform ablation studies to validate the effectiveness of each component of our method.

\begin{figure*}[!t]
    \centering
\includegraphics[width=1.0\linewidth]{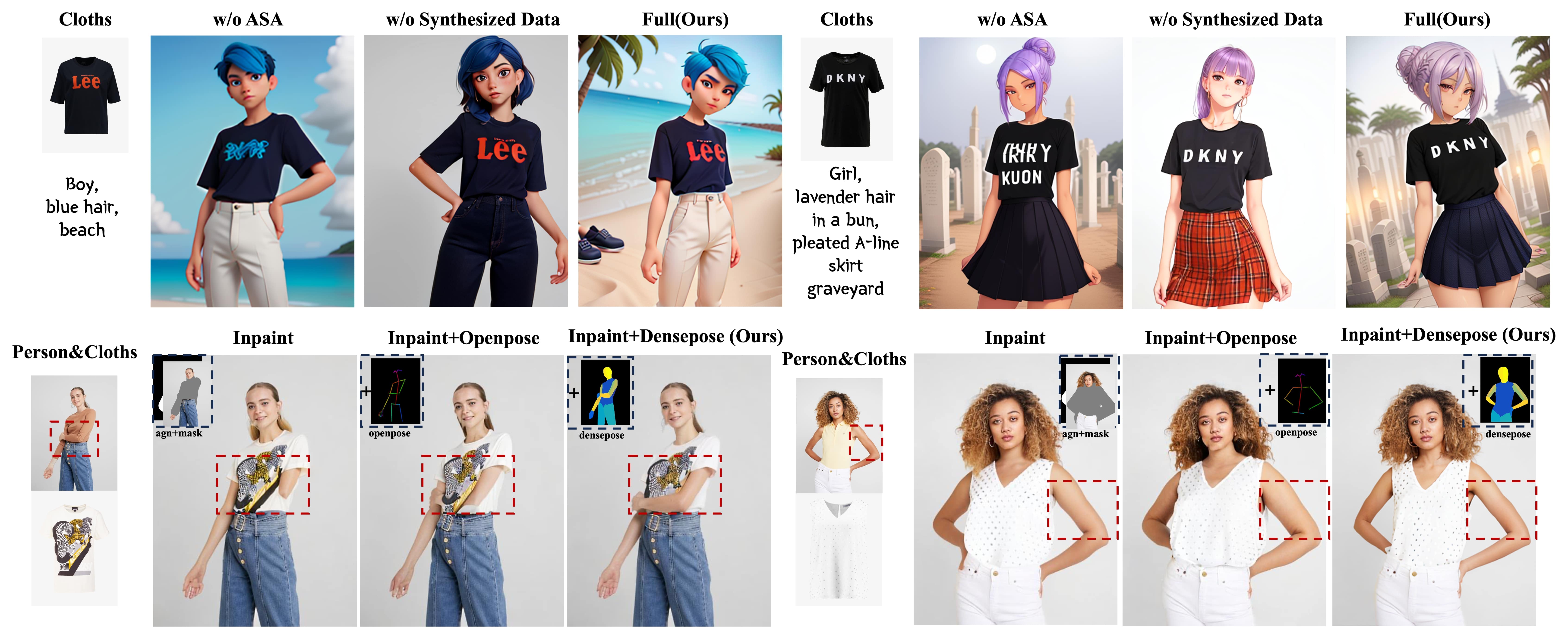}
\vspace{-15pt}
    \caption{Effects of our proposed components. }
    \label{Fig:ab-comp}
\vspace{-0.2cm}
\end{figure*}

\noindent\textbf{Additive Self-Attention.} To study the effectiveness of our ASA operator, we built a model with concatenated self-attention and trained with the same setup. The qualitative and quantitative comparison can be found in Fig. \ref{Fig:ab-comp} and Table \ref{Tab:ASA} . When testing under standard setup, the quantitative performance of both models is comparable. However, when switching the base models into stylized ones, it is visually obvious that the images generated via the CSA model often suffer from unavoidable artifacts. Especially when the text prompt becomes complex, such as a detailed background description, the CSA model often misinterprets the attributes of the garment. Instead, the ASA counterpart generates stylized images with better quality. Both details of the garment and style of the base model can be well preserved by the ASA model.

\begin{table}[!t]
  \small       %此处写字体大小控制命令 
  \centering      %表格居中
  \setlength{\abovecaptionskip}{0cm}
  \caption{Ablation study for garment-centric generation. For User result “a / b”, a is
frequency that each method is chosen as the best result for restoring the clothes, and b represents the best generated result following the text description.}         %表标题
\vspace{-0.2cm}
  \label{Tab:ASA}         %标签，用作引用
	\resizebox{1\linewidth}{!}{% 
\begin{tabular}{l|ccc | ccc | ccc}
  \toprule
    & \multicolumn{3}{c|}{\textbf{SD1.5}} & \multicolumn{3}{c|}{\textbf{Anythingv5~\cite{anything}}} & \multicolumn{3}{c}{\textbf{Disney Pixar~\cite{disney}}} \\ 
    \textbf{Ablation Setups} & CLIP-T↑      & User(\%)↑  & AS↑ & CLIP-T↑      & User(\%)↑  & AS↑ & CLIP-T↑      & User(\%)↑  & AS↑\\
  \midrule
  Ours(w/o ASA)              &  0.328           & 32.57/38.24         & 5.430      
                             &  0.332           & 12.29/25.14         & 5.860       
                             &  0.316           & 10.78/22.30         & 5.631\\
  Ours(w/o Synthesized data) &  0.254           & 35.04/28.71         & 5.107       
                             &  0.275           & 21.83/14.86         & 5.625
                             &  0.280           & 24.71/12.99         & 6.02 \\
  Ours(full)                 &  0.327           & 32.38/33.04         & 5.598       
                             &  0.321           & 65.87/60.00         & 5.919
                             &  0.314           & 64.50/64.71         & 5.865\\
  \bottomrule
\end{tabular}}
\vspace{-0.3cm}
\end{table}

\noindent\textbf{Effects of Synthesized Data.} We also ablate the benefits of using synthesized data. The comparing setups are: 1) training with VITON-HD and 2) training with our synthesized data. From Table \ref{Tab:ASA}, we found that on the GC genearation benchmark, their performances are close under the standard settings. However, when testing the prompt following capacity on other stylized base mdoels, the model trained with synthesized data significantly outperforms its counterpart. From Fig. \ref{Fig:ab-comp}, the images generated by the model trained without synthesized data would display monotonous backgrounds, which reflects a characteristic of the VITON-HD dataset. 

\begin{table}[!tb]
  \small       %此处写字体大小控制命令 
  \centering      %表格居中
  \setlength{\abovecaptionskip}{0cm}
  \caption{Ablation study for try-on ControlNet.}         %表标题
  \label{Tab:ab-quan-tab}         %标签，用作引用
  \vspace{-0.1cm}
\begin{tabular}{l|ccccc}
  \toprule
  Ablation Setups            & LPIPS↓   & SSIM↑   & FID↓  & KID↓  & DINO-M↑ \\
  \midrule
  Inpainting                 &  0.097       & 0.852         & 9.11     & 0.303         & 0.661        \\
  +OpenPose                  &  0.080       & 0.870         & 8.60     & 0.251         & 0.663        \\
  +DensePose(Ours)           &  0.077      & 0.877         & 8.02     & 0.110         & 0.661    \\
  \bottomrule
\end{tabular}
\vspace{-0.4cm}
\end{table}

\noindent\textbf{Comparison with Different Try-on ControlNet.} We evaluate three types of try-on ControlNet, with three different control inputs: (1) garment-agnostic input and mask, (2) garment-agnostic input, mask and OpenPose, and (3) garment-agnostic input, mask, and DensePose. Fig.~\ref{Fig:ab-comp} shows a visual comparison of these three different setups of try-on ControlNets. All of these ControlNets enable to preserve detailed clothing texture due to our proposed garment encoder. However, without the aid of geometric information, it is difficult for our model to generate accurate human body, including arms and hands. In comparison to OpenPose control, DensePose control offers more geometric information. As shown in Fig.~\ref{Fig:ab-comp}, replacing OpenPose with DensePose as a condition results in the disappearance of artifacts caused by occulusion in the generated images, with the thickness of the arms also better matching source model. We also conduct a comprehensive quantitative experiment to verify the effectiveness of our proposed DensePose-based try-on ControlNet. As presented in Table~\ref{Tab:ab-quan-tab}, when adding the OpenPose condition on top of the base inpainting ControlNet, all metrics improve, as the model gains more information about the human body structure. Replacing OpenPose with DensePose leads to improvements in the detailed morphology of the human body, and all metrics also show enhancements.
%\noindent \textbf{Reference Posing Branch}. 

% \subsubsection{Effectiveness of Finetuning} We also replace the finetuned controlnets with pretrained ones, to verify if the finetuning is necessary. 

% \noindent \textbf{}. 

\section{Conclusion}
In this paper, we propose our novel unified framework, StableGarment, to solve garment-driven generation tasks, including garment-centric text-to-image, garment-centric controllable text-to-image, garment-centric stylized image generation, and virtual try-on with flexible inputs.
We introduce a garment encoder to capture detailed clothing features, which consists of a trainable copy of UNet with designed additive self-attention (ASA) layers. These designs help our encoder capture detailed clothing features and switch between stylized base models. Furthermore, to preserve the model's ability to follow prompts, we propose a novel data engine to generate high-quality synthesized data. We also present a carefully designed try-on ControlNet to address the classic virtual try-on task. Extensive experiments have demonstrated that our approach delivers state-of-the-art (SOTA) results among existing virtual try-on methods and exhibits high flexibility with broad potential applications in various related fields.

% maybe do not need
% \input{chapters/6-limitation}
\clearpage  
\bibliographystyle{splncs04}
\bibliography{main}

\clearpage

\appendix
\section*{Supplementary}
% \section{Supplementary}

\renewcommand\thesection{\Alph{section}}
In this supplementary material, we have provided additional details and experiments to support our main paper. We have introduced our data engine and implementation details in Section~\ref{dataengine} and Section~\ref{implementationdetail}, respectively. In Section ~\ref{quantitativeexp}, we have presented additional quantitative experiments, including the evaluation of the cross-dataset analysis between VITON-HD and Dress Code training datasets. We have also introduced our garment benchmark in Section~\ref{benchmark}. In Section~\ref{userstudy}, we have provided more information about our user study, including metric computation and query format. We have discussed more applications in Section~\ref{Application} and the limitations of our method in Section~\ref{limitations}. 

\section{Data Engine}
\label{dataengine}

\begin{figure*}[!h]
    \centering
\includegraphics[width=1.0\linewidth]{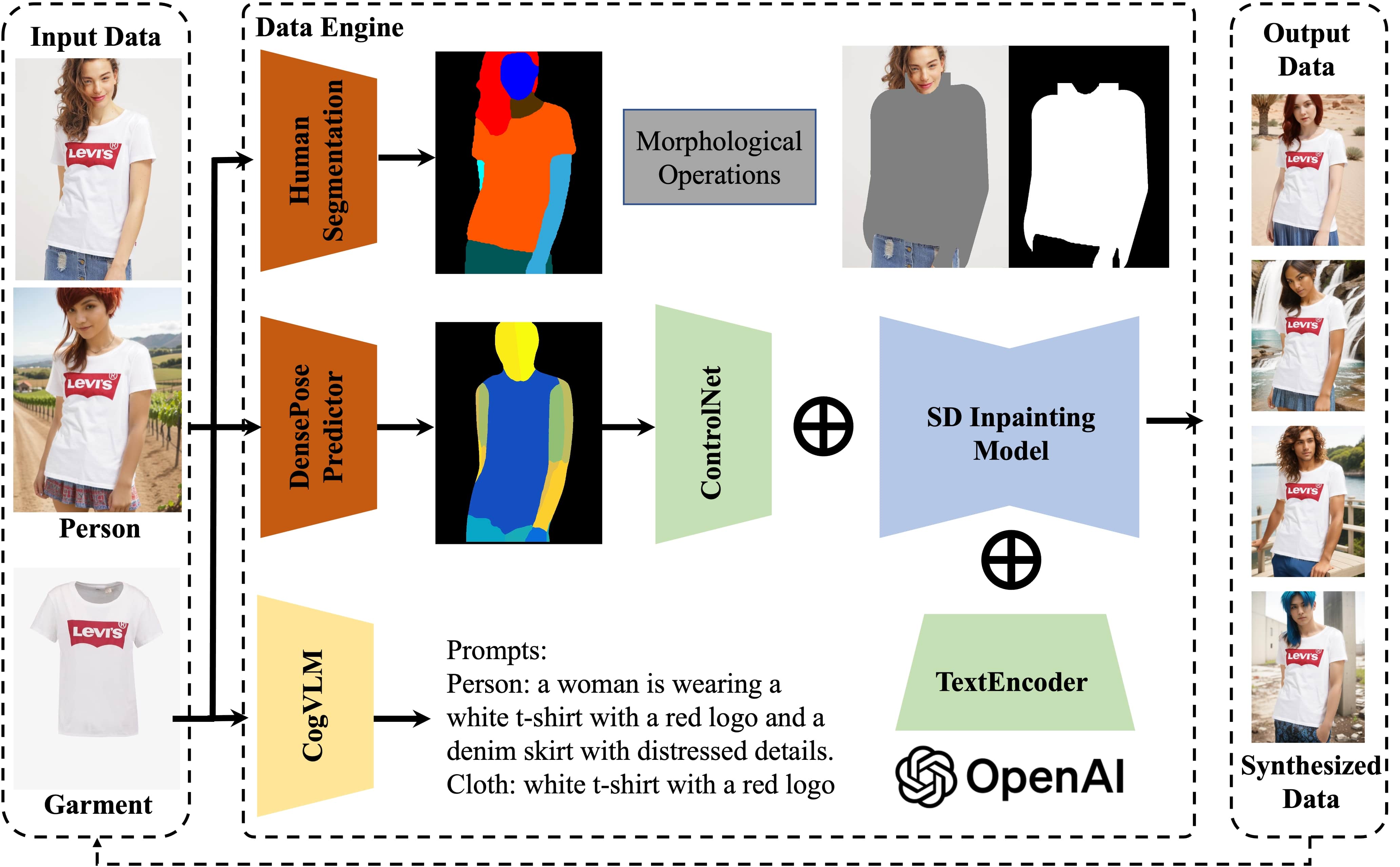}
\vspace{-0.6cm}
    \caption{Data engine framework. }
    \label{fig:data_engine}
\vspace{-0.6cm}
\end{figure*}

The overall framework for our data engine are as shown in Fig.~\ref{fig:data_engine}.
The complete workflow is as follows:
\begin{enumerate}
    \item Adopt human segmentation model and DensePose prediction model~\cite{detectron2} to get parsing map and dense map.
    \item Combine predicted maps with morphological operations to estimate the garmnet-agnostic mask and corresponding image content.
    \item Utilize GPT4 to generate inpaint prompts and employ the DensePose ControlNet~\cite{DensePoseControlNet} and Stable Diffusion inpainting model~\cite{epicinapint} to inpaint the sythesized outputs.
    \item Use CogVLM~\cite{cogvlm} as a tagger to label each image with a detailed description. 
    \item Repeat steps 1-4 to generate satisfactory synthesized outputs.
\end{enumerate}

\section{Implementation Details}
\label{implementationdetail}

\begin{figure*}[!h]
    \centering
\includegraphics[width=0.65\linewidth]{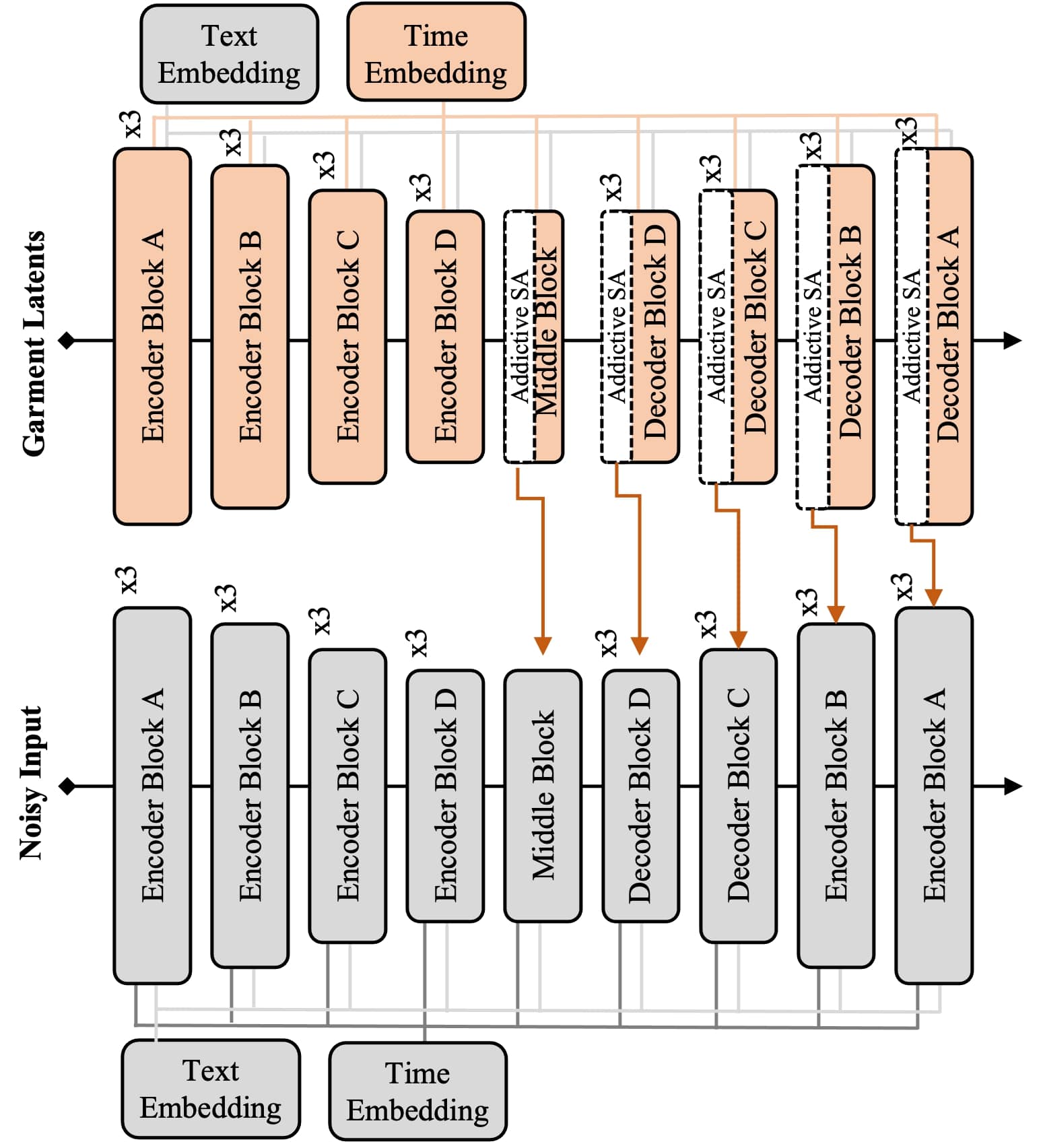}
    \caption{The framework of our proposed garment encoder. }
    \label{fig:garment_encoder}
\vspace{-0.6cm}
\end{figure*}

For subject-driven generation, we employed Stable Diffusion V1-5 as the pre-trained diffusion model and trained it on our synthesized dataset. Our synthesized dataset consists of 11,436 images with corresponding DensePose, text description and parsing images. We adopt the flip operation for data augmentation. The model underwent 50k iterations of training on 8 H800 GPUs, with a batch size of 8 per GPU and a learning rate of 1e-4. Inference was performed using DDIM as the sampler, with a step size of 25 and a guidance scale set to 3.
In the context of virtual try-on training, we follow the previous setup at the first stage and only train our garment encoder on official dataset. At the second stage, we frozen our garment encoder and only train our DensePose try-on ControlNet. We adopt filp, random shift and random scale for data augmentation. We initialize our ControlNet with the DensePose ControlNet~\cite{DensePoseControlNet}. We train our second stage for 20k interations on 8 H800 GPUs, with a batch size of 8 per GPU and a learning rate of 1e-5. Inference was performed using DDIM as the sampler with inapinting pipeline, with a step size of 25 and a guidance scale set to 3. The detail of our garment encoder is shown in Fig.~\ref{fig:garment_encoder}.

\section{Additional Quantitative Experiments}
\label{quantitativeexp}

\begin{table}[!h]
  \small       %此处写字体大小控制命令 
  \centering      %表格居中
  \setlength{\abovecaptionskip}{0cm}
  \caption{Quantitative comparison on cross data setting. The best and the second best results are denoted as \textbf{Bold} and \underline{underline}, respectively.}         %表标题
  \label{Tab:cross}
  \vspace{-0.2cm}
\scalebox{0.95}{%标签，用作引用
\begin{tabular}{l|cccc}
  \toprule
  \textbf{Model} & \textbf{LPIPS}↓  & \textbf{SSIM}↑ & \textbf{FID}↓ &\textbf{KID}↓  \\
  \midrule
  VITON-HD~\cite{viton-hd}  & 0.187  & 0.853 & 44.26 & 2.882  \\
  HR-VTON~\cite{hr-viton}   &  0.108  & 0.909 & 19.97 & 0.735  \\
  LADI-VTON~\cite{ladi-vton}&  0.101  & 0.901 & 16.34 & 0.536  \\
  DCI-VTON$^*$~\cite{dci-vton}  &  0.124  & 0.898 & 18.81 & 0.802  \\ 
  StableVITON$^*$~\cite{stableviton}   &  \underline{0.060}  & \underline{0.911} & \underline{12.58} & \underline{0.170}  \\
  GP-VTON$^*$~\cite{gp-vton}&  0.385  & 0.887 & 65.71 & 6.601  \\
  Paint-by-Example~\cite{paint-by-example}   &  0.087  & 0.889 & 14.17 & 0.478  \\
  Ours   &  \textbf{0.046}  & \textbf{0.944} & \textbf{11.15} & \textbf{0.063}  \\
  \bottomrule
\end{tabular}}
\vspace{-0.2cm}
\end{table}

In this section, we conduct a cross-dataset analysis between the VITON-HD~\cite{viton-hd} and Dress Code~\cite{dresscode} training datasets. We report the results in Table~\ref{Tab:cross}, presenting only the metrics measured by previous methods. The table demonstrates the effectiveness of our proposed method, as we outperform other methods across all metrics. Furthermore, the qualitative results presented below verify the remarkable capacity of our approach in adapting to different garment domains while preserving intricate garment details.

\section{Garment Benchmark}
\label{benchmark}
We have developed a garment-centric generation benchmark specifically tailored for evaluating subject-driven generation methods, serving as a comparative platform against our own methodology. Primarily honing our model for the try-on task, we meticulously curated this benchmark from a subset of the VITON-HD test set. Comprising 14 distinct garments and 6 meticulously crafted text prompts, our benchmark offers a comprehensive evaluation framework. To ensure a diverse representation, we selected garments based on various attributes including color, shape, pattern, size, and more, although not exhaustive in its coverage. These prompts, generated using GPT-4 with a consistent format, mirror the essence of try-on tasks by focusing on the individual's appearance while incorporating additional attributes such as skin tone and hairstyle.

\begin{figure*}[!t]
    \centering
\includegraphics[width=1.0\linewidth]{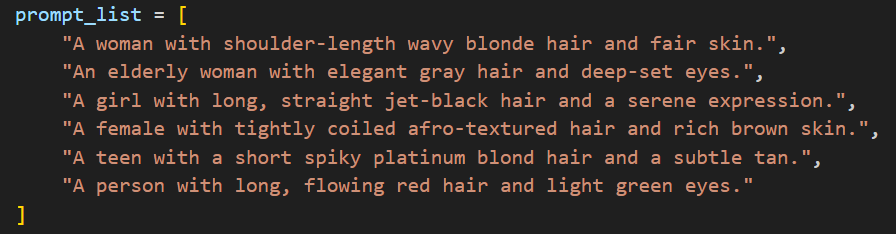}
\vspace{-15pt}
    \caption{Evaluation prompts for subject-driven generation methods. }
    \label{Fig:prompt_list}
\vspace{-0.6cm}
\end{figure*}

\begin{figure*}[!t]
    \centering
\includegraphics[width=1.0\linewidth]{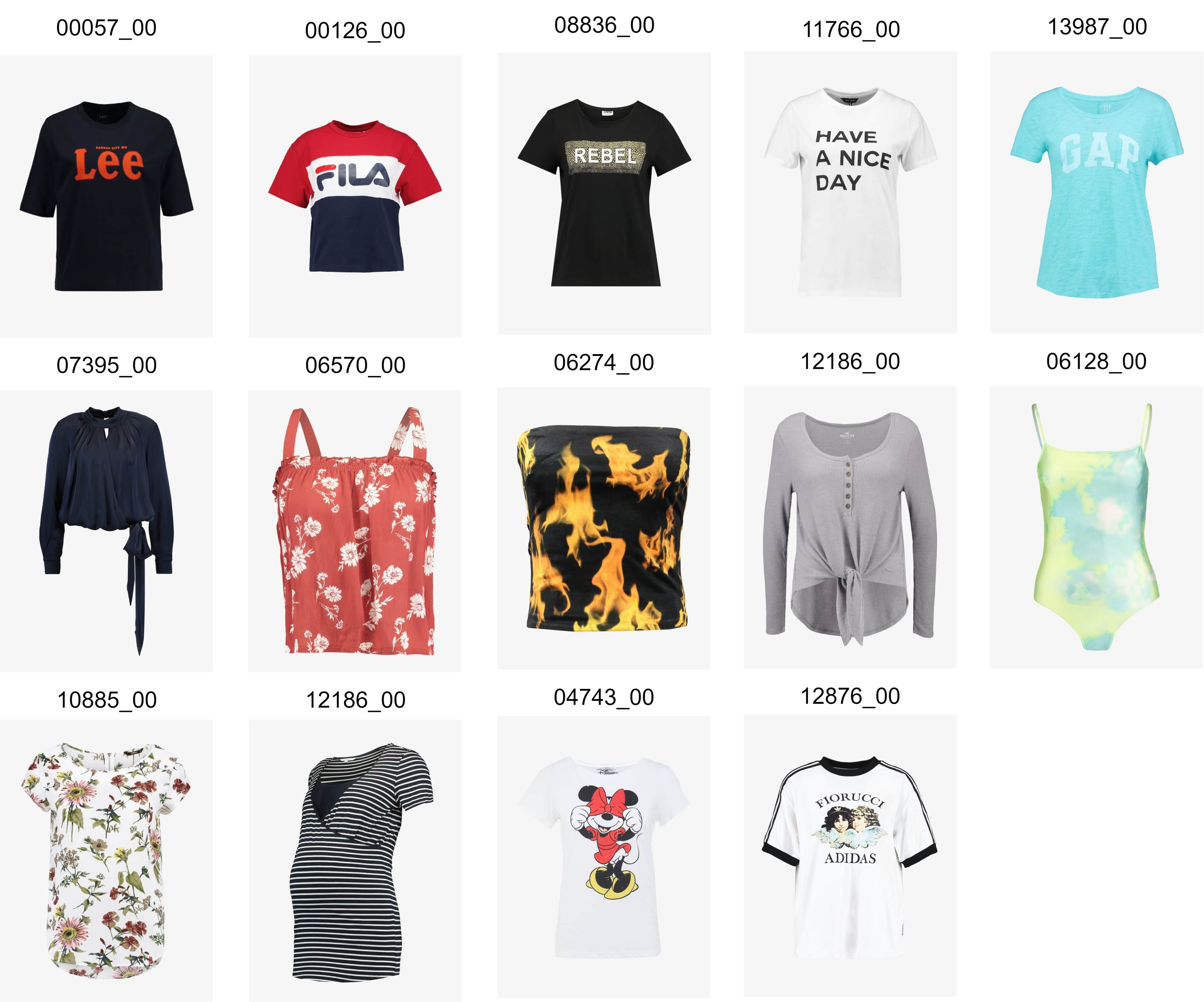}
\vspace{-15pt}
    \caption{The selected garments for evaluating subject-driven generation methods. }
    \label{Fig:garments}
\vspace{-0.6cm}
\end{figure*}

The generated prompts and selected garments with corresponding ids are shown in the Fig.\ref{Fig:prompt_list} and Fig.\ref{Fig:garments}, respectively.

% \begin{figure*}[!t]
%     \centering
%   \begin{subfigure}{0.5\linewidth}
%     \includegraphics[width=0.98\linewidth]{images/prompt_list.png}
%     \caption{The generated prompts.}
%     \label{Fig:prompts}
%   \end{subfigure}%
%   \begin{subfigure}{0.5\linewidth}
%     \includegraphics[width=0.98\linewidth]{images/subject-driven_benchmark.png}
%     \caption{The selected garments.}
%     \label{Fig:garments}
%   \end{subfigure}%
%   \label{Fig:garment_benchmark}
% \end{figure*}

We compare our methods with three subject-driven methods, BLIP-Diffusion, IP-Adapter and ELITE. We take official implementation of IP-Adapter and ELITE and use BLIP-Diffusion implementation by diffusers. We set all parameters following examples provided by implementation, respectively.

\section{User Study}
\label{userstudy}
\begin{figure*}[!t]
    \centering
\includegraphics[width=1.0\linewidth]{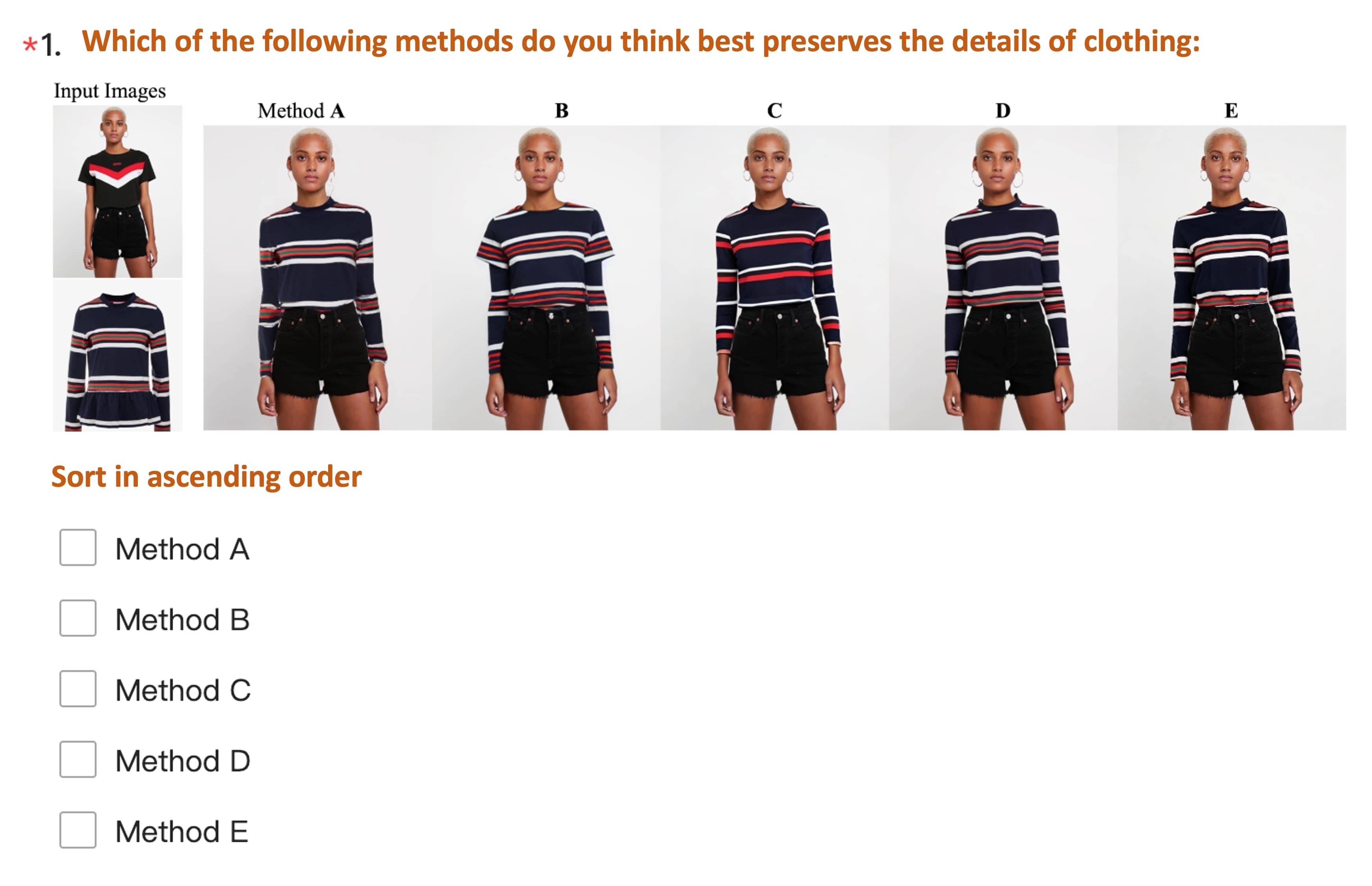}
\vspace{-15pt}
    \caption{Query format.}
    \label{fig:query}
\vspace{-0.6cm}
\end{figure*}

\noindent\textbf{Metric Computation.}
In our user study to assess the effectiveness of virtual try-on tasks, we utilize two principal metrics: human preference and human scores. Participants are requested to rank the results from our model alongside those from comparative baseline models, focusing on three key aspects: garment identity, try-on quality, and the garment-agnostic preservation. Human preference is measured by the frequency at which each method is ranked as the preferred choice relative to the baselines. Human scores, on the other hand, are derived from a weighted scheme reflecting the participants' rankings. The formula for computing the human scores is as follows:
\begin{equation}
S = \frac{\sum(f \times W)}{N},
\end{equation}
where $S$ refers to the average comprehensive score of an option, $f$ refers to the preference frequency, $N$ refers to the total number of responses for the item, and $W$ refers to the inverse rank assigned by a participant, multiplied by the total count of methods under comparison. This weighting system is designed to proportionally recognize the preferences indicated by participants. The format of our used query is as shown in Fig.~\ref{fig:query}.

\section{More Applications}
\label{Application}
\begin{figure*}[!t]
    \centering
\includegraphics[width=1.0\linewidth]{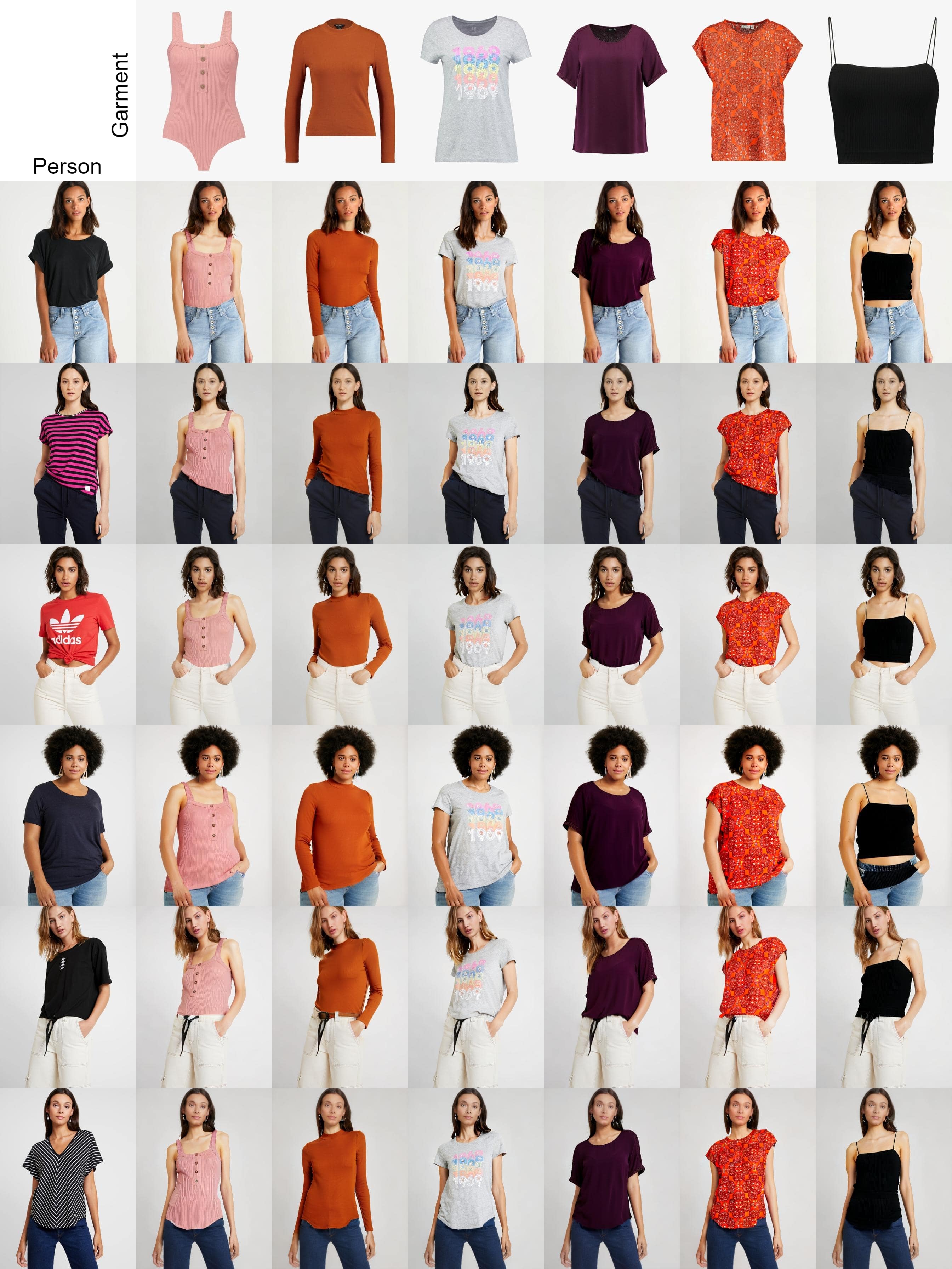}
\vspace{-15pt}
    \caption{Our try-on results on VITON-HD dataset at 512x384 resolution. }
    \label{Fig:add_quality-1}
\vspace{-0.6cm}
\end{figure*}

\begin{figure*}[!t]
    \centering
\includegraphics[width=1.0\linewidth]{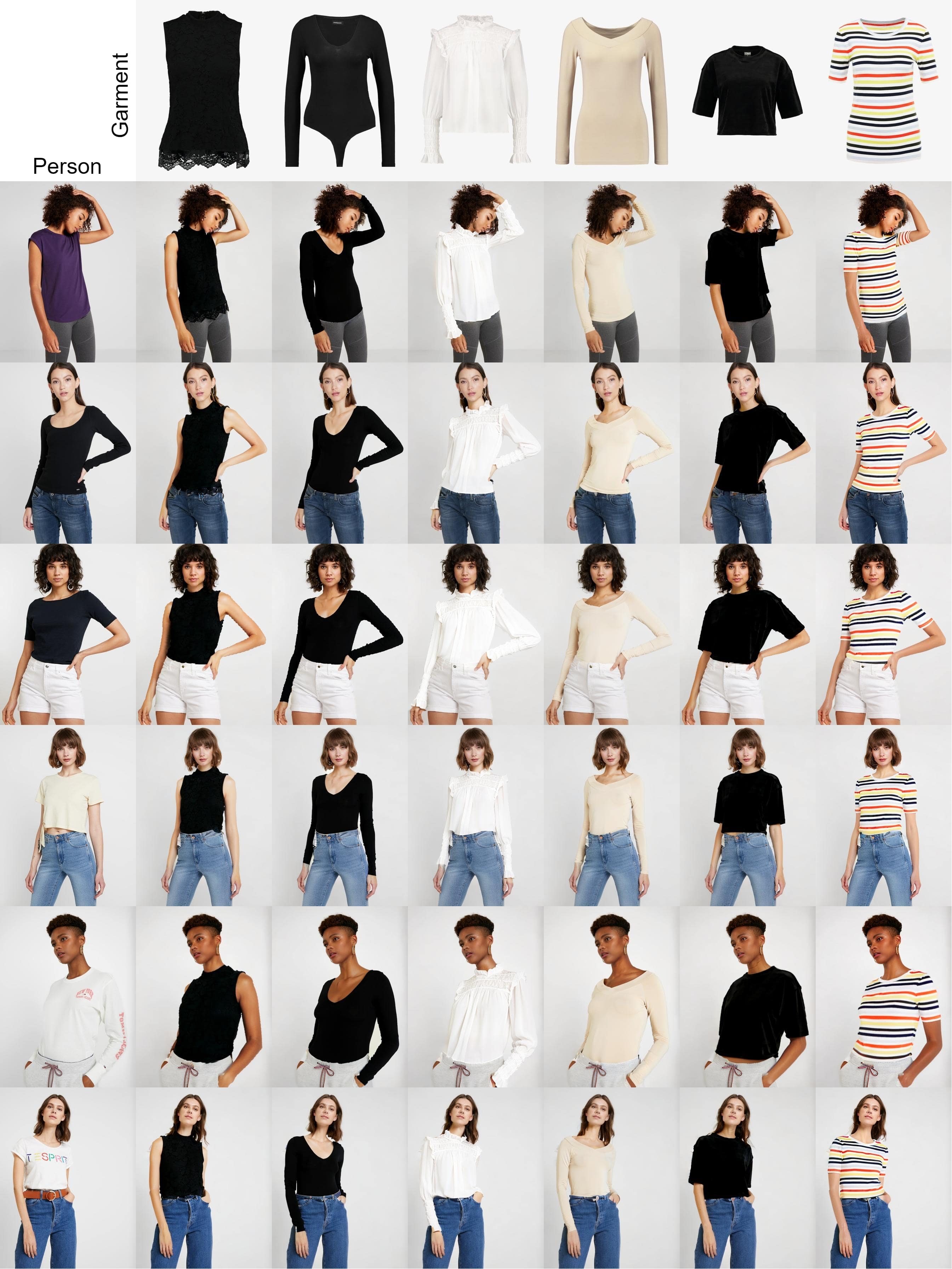}
\vspace{-15pt}
    \caption{Our try-on results on VITON-HD dataset at 512x384 resolution. }
    \label{Fig:add_quality-2}
\vspace{-0.6cm}
\end{figure*}

\noindent\textbf{Virtual try-on task.} We present the results generated by our StableGarment model in Figures \ref{Fig:add_quality-1} and \ref{Fig:add_quality-2}. Our model can learn accurate warping transformations while preserving the intricate details of the garments.

\begin{figure*}[!h]
    \centering
\includegraphics[width=1.0\linewidth]{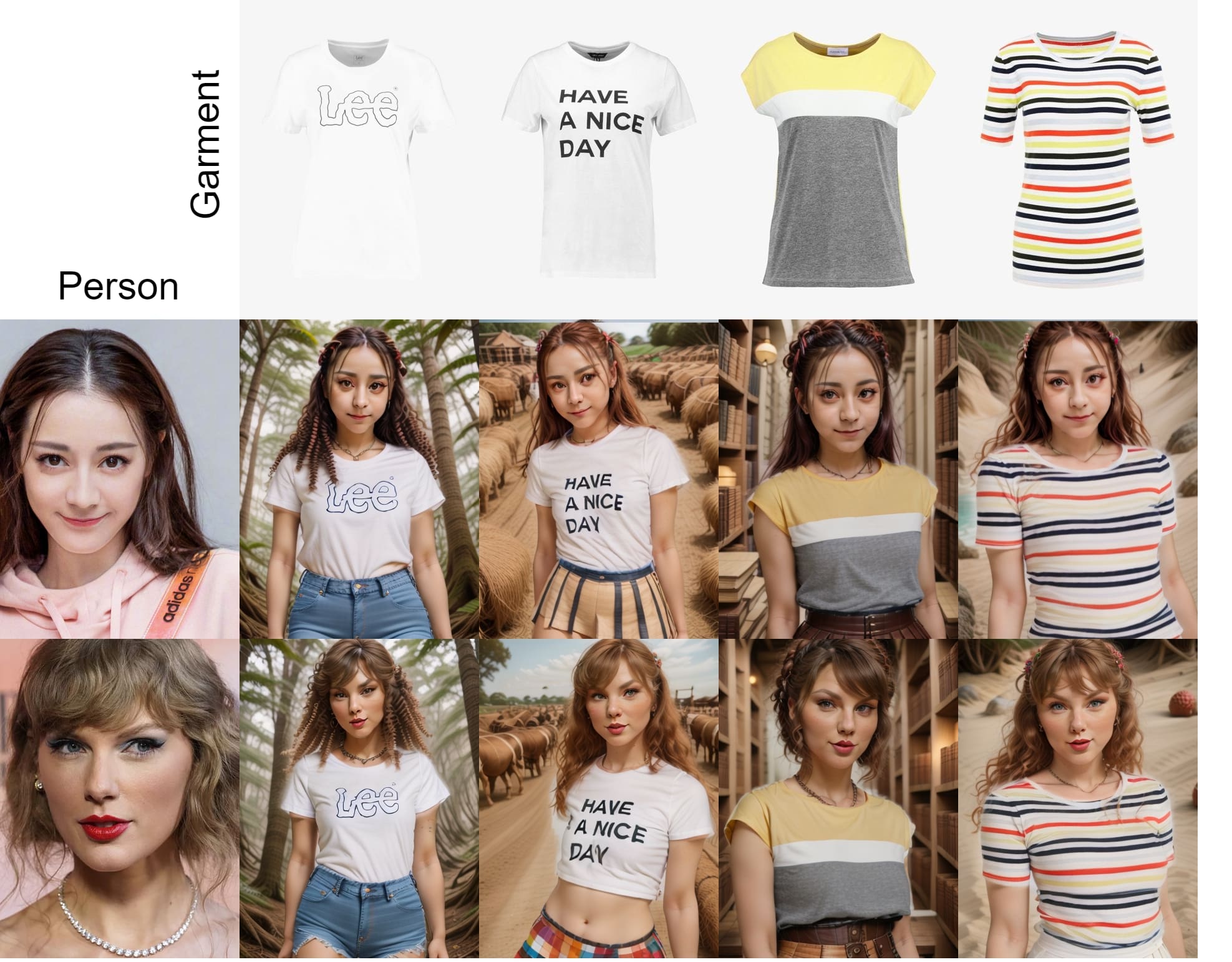}
\vspace{-15pt}
    \caption{Our text-to-image try-on results integrated with IP-Adapter.}
    \label{Fig:gcg-1}
\vspace{-0.6cm}
\end{figure*}

\noindent\textbf{Combined with IP-Adapter~\cite{ip-adapter}.} Our model, when combined with the IP-Adapter~\cite{ip-adapter}, enables the generation of target individuals wearing target garments. We leverage the ID preservation capability of IP-Adapter FaceID-PlusV2~\cite{IP-Adapter-FaceID} to deliver an authentic try-on experience. The visual results, shown in Figure~\ref{Fig:gcg-1}, demonstrate the remarkable compatibility of our methods with current plugins.

\begin{figure*}[!t]
    \centering
\includegraphics[width=1.0\linewidth]{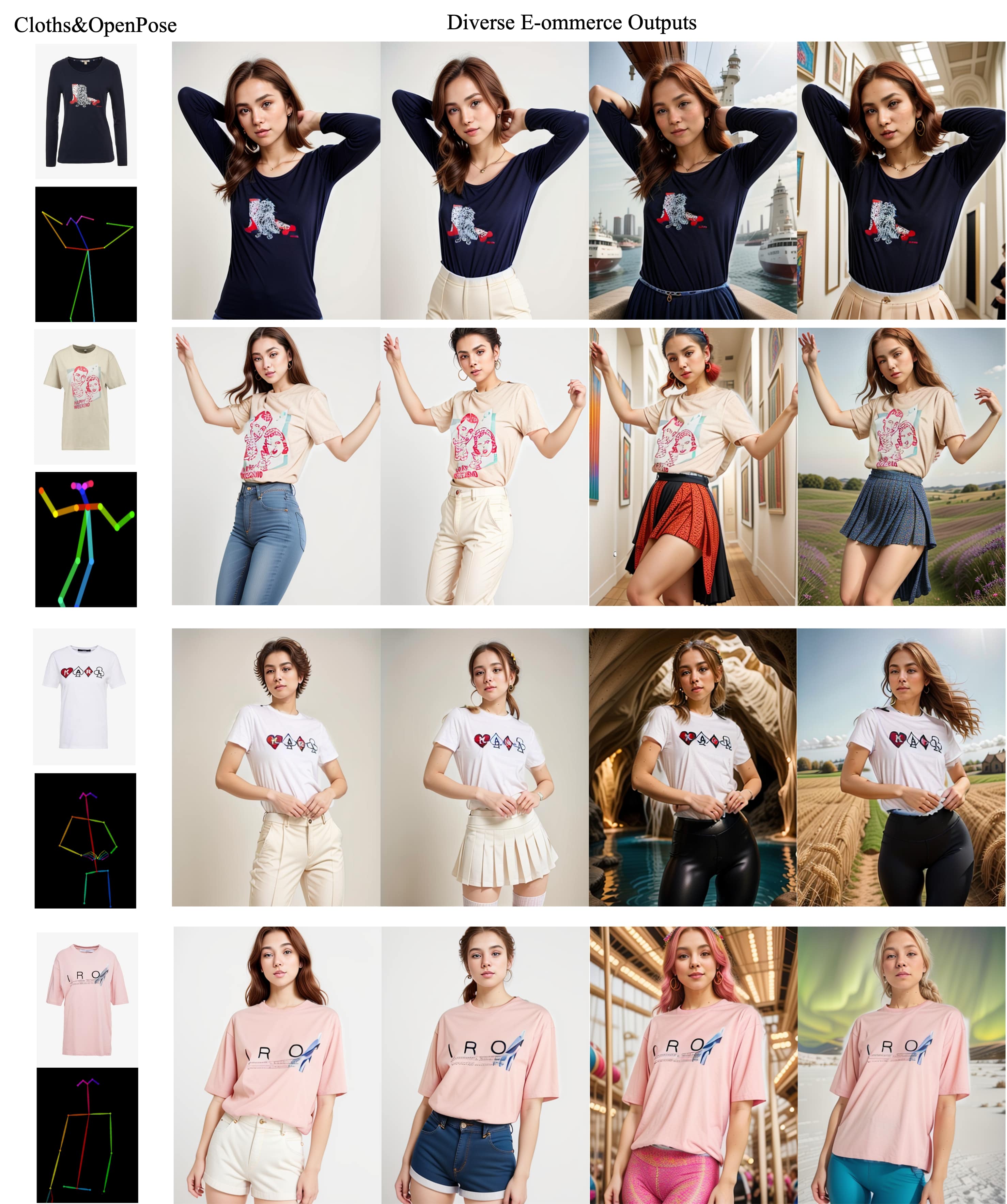}
\vspace{-15pt}
    \caption{E-commerce model generation.}
    \label{fig:e-commerce}
\vspace{-0.6cm}
\end{figure*}

\noindent\textbf{E-commerce model generation.} Leveraging the capabilities of ControlNet~\cite{controlnet}, our model can generate e-commerce models guided by specific conditions, such as OpenPose and DensePose. We present the OpenPose-guided generation results in Figure~\ref{fig:e-commerce}.

\begin{figure*}[!t]
    \centering
\includegraphics[width=1.0\linewidth]{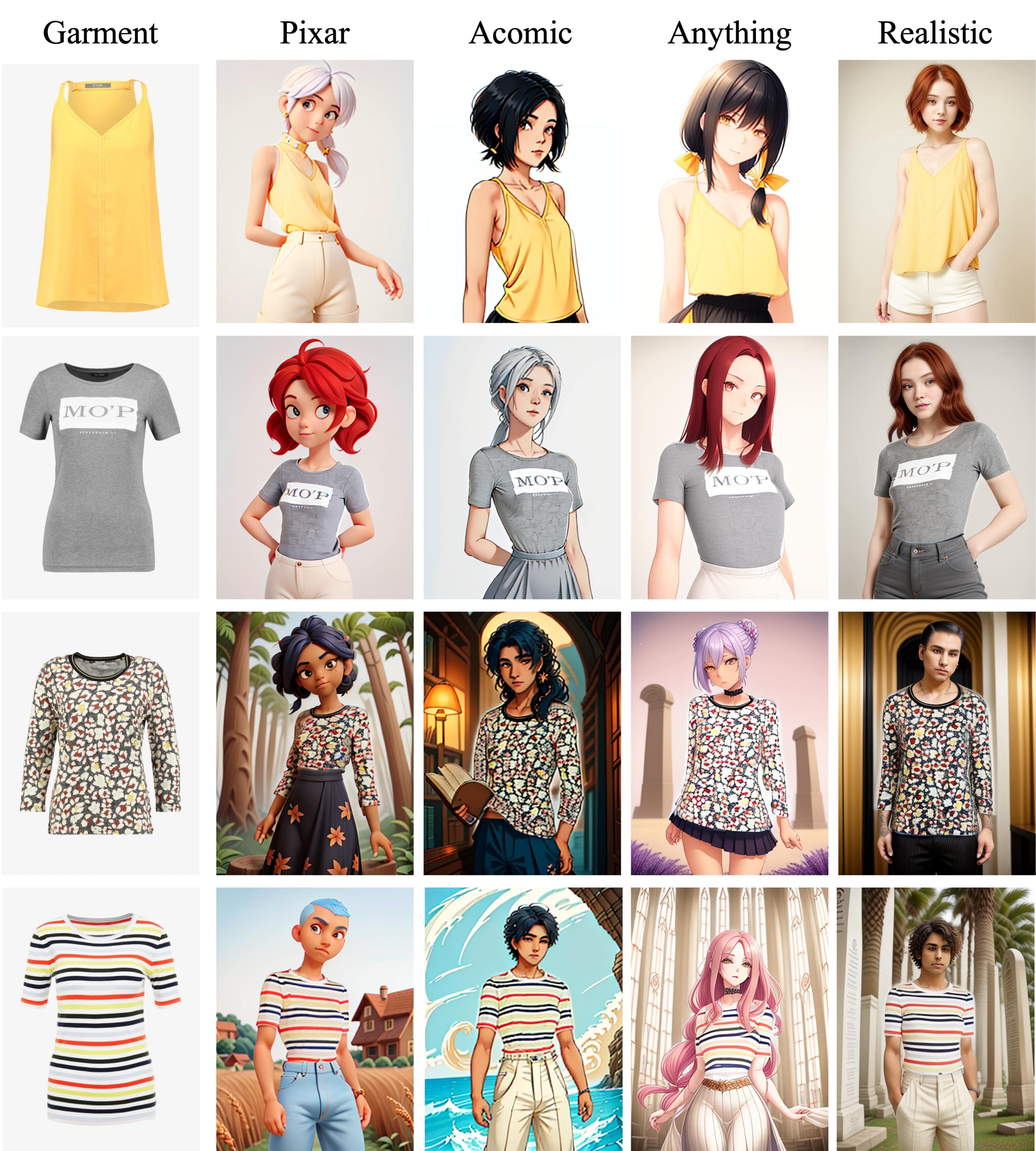}
\vspace{-15pt}
    \caption{Stylized garment-centric generation.}
    \label{fig:stylized-garment-centric-generation}
\vspace{-0.6cm}
\end{figure*}

\noindent\textbf{Stylized garment-centric generation.} Furthermore, by replacing the standard Stable Diffusion 1.5 model with other diverse base models, we can generate creative and stylized outputs while preserving the intricate details of the garments. We present these results in Figure~\ref{fig:stylized-garment-centric-generation}.

% \noindent\textbf{Real scene virtual try-on.}

\section{Limitations and Discussion}
\label{limitations}

% VAE problem for bad text reconstruction.

\begin{figure*}[!t]
    \centering
\includegraphics[width=1.0\linewidth]{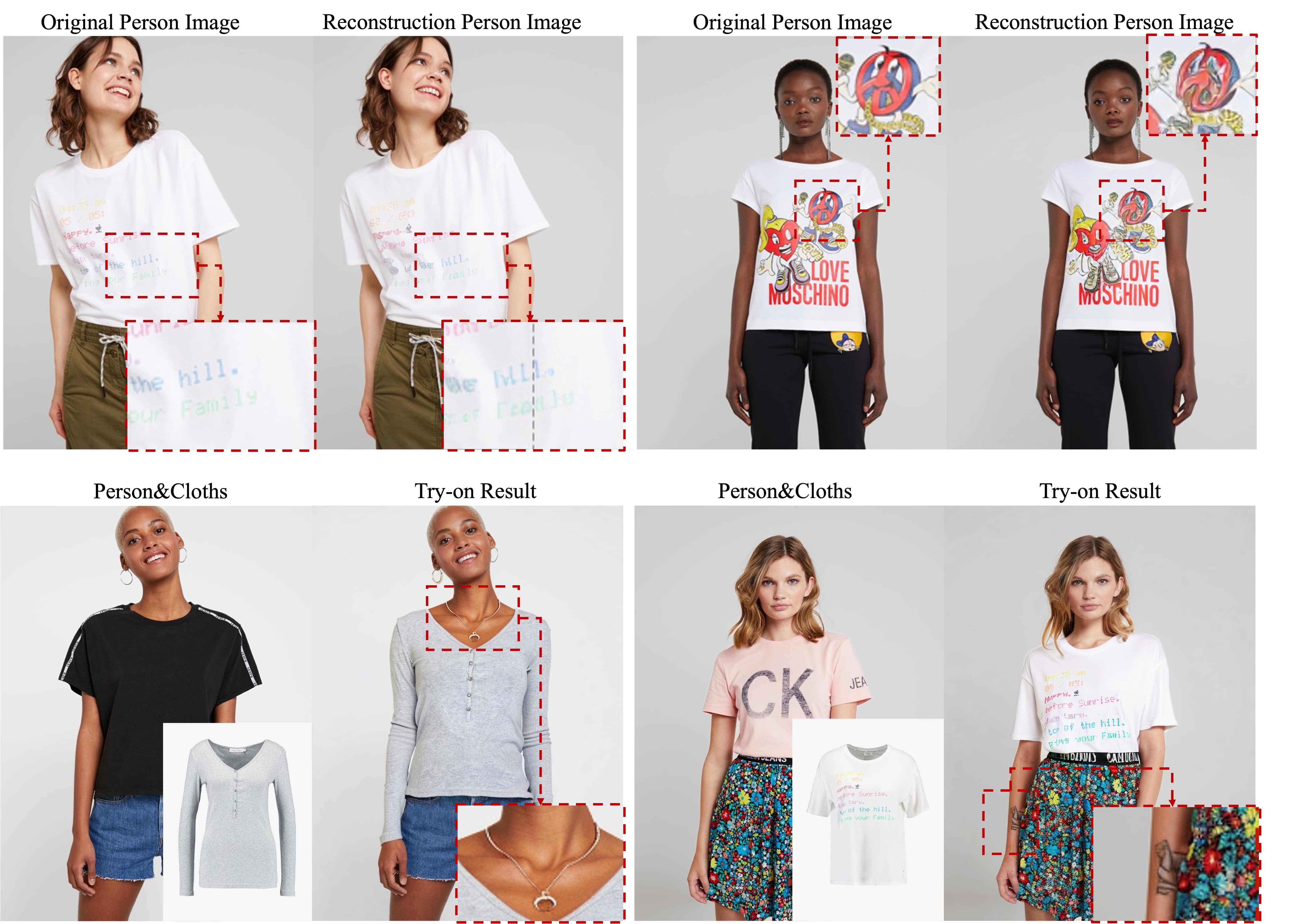}
\vspace{-15pt}
    \caption{Bad cases. }
    \label{fig:lim}
\vspace{-0.6cm}
\end{figure*}

% Wrong accessories generation, e.g. necklace, or ring
Our model faces two main challenges: the VAE reconstruction problem and the generation of incorrect accessories.
Fig.\ref{fig:lim} shows that the standard VAE used in Stable Diffusion 1.5 fails to preserve all the details of the garment through a simple reconstruction. Although we tried to employ a more advanced version of VAE\cite{MSE_VAE}, the problem cannot be fully solved. Therefore, research on how to better preserve detailed information may need further improvement.
Meanwhile, in the context of incorrect accessory generation, we found that the trained model may tend to generate some incorrect accessories during inference, especially for the high-resolution version. This is often caused by inaccurate parsing conditions, e.g., garment-agnostic masks or DensePose. We leave this problem for future work.

% \input{chapters/supp}

\end{document}